\documentclass[runningheads]{llncs}

\usepackage{graphicx} % Required for inserting images
\usepackage{changepage}
\usepackage{graphicx}
\usepackage{tabularx}
\usepackage{xcolor}
\usepackage{colortbl}
\usepackage{float}
\usepackage{newfloat}
\usepackage{enumitem}

\definecolor{Gray}{gray}{0.90}
\definecolor{VeryLightGray}{gray}{0.98}
\definecolor{LightCyan}{rgb}{0.88,1,1}
\usepackage{transparent}
\definecolor{LightPink}{rgb}{1.0, 0.83, 0.88}
\colorlet{LightPink}{LightPink!50} % % opacity
\definecolor{LightBlue}{rgb}{0.6, 0.8, 1.0}
\definecolor{DarkBlue}{rgb}{0.05, 0.30, 0.85}
\definecolor{DarkPink}{rgb}{0.80, 0.20, 0.40}

\usepackage[labelfont=bf]{caption}
\usepackage{subcaption}
\usepackage{afterpage}
\usepackage{pdfpages}
\newcolumntype{a}{>{\columncolor{Gray}}c}
\newcolumntype{b}{>{\columncolor{VeryLightGray}}c}
\usepackage{makecell}

\usepackage{amsmath}
\usepackage{multirow} % for the tables
\usepackage[utf8]{inputenc}
\usepackage{algorithm}
\usepackage{algpseudocode}

\definecolor{cust}{RGB}{84,178,255} 
\usepackage[colorlinks,allcolors=cust]{hyperref}
\makeatletter
\newcommand{\printfnsymbol}[1]{%
  \textsuperscript{\@fnsymbol{#1}}%
}
\makeatother

\usepackage{multicol}
\usepackage{needspace}

\usepackage{xr}
\usepackage{etoc}

\usepackage[final, authormarkup=none]{changes}   % "final" hides markup; "draft" shows it
\definechangesauthor[name={Lucas}, color=DarkBlue]{lu}   % temp IDs 
\definechangesauthor[name={Johannes}, color=DarkBlue]{jo}

\usepackage[a4paper, margin=1.5in]{geometry} % Added to control margins

\begin{document}
\newgeometry{left=0.85in, right=0.85in, top=1.15in} % Temporarily tighter margins for title/authors

\title{Histo-Miner: Deep learning based tissue features extraction pipeline from H\&E whole slide images of cutaneous squamous \\  cell carcinoma}
% \date{April 2025}
\titlerunning{Histo-Miner}

\author{Lucas Sancéré \inst{1,2,3} \and Carina Lorenz\inst{3,5,6} \and Doris Helbig \inst{7}  \and Oana-Diana Persa \inst{8} \and Sonja Dengler \inst{9} \and Alexander Kreuter \inst{10} \and Martim Laimer \inst{11} \and \added[id=lu]{Roland Lang \inst{11} } \and Anne Fröhlich \inst{12} \and Jennifer Landsberg  \inst{12} \and Johannes Brägelmann\inst{3,5,6, \added[id=lu]{13}} \thanks{Equal contribution} \and Katarzyna Bozek \inst{2,3,4} \printfnsymbol{1}  } 
\authorrunning{Sancéré et al.}

\institute{
\fontsize{9}{10}\selectfont
    \noindent Faculty of Mathematics and Natural Sciences, University of Cologne, Cologne, North Rhine-Westphalia, Germany \\ %1
    \and
    Institute for Biomedical Informatics, Faculty of Medicine and University Hospital Cologne, University of Cologne, Cologne, North Rhine-Westphalia, Germany \\ %2
    \and
    Center for Molecular Medicine Cologne (CMMC), Faculty of Medicine and University Hospital Cologne, University of Cologne, Cologne, North Rhine-Westphalia, Germany \\ %3
    \and
    Excellence Cluster on Cellular Stress Responses in Aging-Associated Diseases (CECAD), University of  Cologne, Cologne, North Rhine-Westphalia, Germany \\ %4
    \and
    University of Cologne, Faculty of Medicine and University Hospital Cologne, Department of Translational Genomics, Cologne, Germany \\ %5
    \and
    University of Cologne, Faculty of Medicine and University Hospital Cologne, Mildred Scheel School of Oncology, Cologne, Germany \\%6
    \and
    Department for Dermatology, University Hospital Cologne, Cologne, Germany \\%7
    \and
    Department of Dermatology and Allergy, School of Medicine, Technical University of Munich, Bavarian Cancer Research Center (BZKF), Munich, Germany \\%7
    \and
    Department of Dermatology, Dortmund Hospital gGmbH, University Witten/Herdecke, 44137 Dortmund, Germany \\%7
    \and
    Department of Dermatology, Venereology and Allergology, Helios St. Elisabeth Hospital Oberhausen, University Witten/Herdecke, Oberhausen, Germany\\%7
    \and
    Department of Dermatology and Allergology, University Hospital of the Paracelsus Medical University Salzburg, Salzburg, Austria \\%7
    \and
    Department of Dermatology and Allergology, University Hospital Bonn, Bonn, Germany
    \and
    \added[id=lu]{Medical Clinic III for Oncology, Hematology, Immune-Oncology and Rheumatology, University Hospital Bonn (UKB), Germany} \\
    \email{lsancere@uni-koeln.de, johannes.braegelmann@uni-koeln.de, k.bozek@uni-koeln.de}
}

\maketitle              % typeset the header of the contribution

\begin{abstract}
%\fontsize{9pt}{10pt}\selectfont
%\fontsize{10pt}{11pt}\selectfont
Recent advances in digital pathology have enabled comprehensive analyses of Whole-Slide Images (WSIs) from tissue samples, leveraging high-resolution microscopy and computational capabilities. Despite this progress, available tools for automatic cell type identification perform poorly on skin tissue, e.g. in the classification of non-melanoma tumor cells. This is due to a paucity of labeled training data sets and high morphological similarities between tumor and non-tumor epithelial cells in the skin. Here, we propose Histo-Miner, a deep learning-based pipeline designed for the analysis of skin WSIs. To this end we generated two new datasets using WSIs of cutaneous Squamous Cell Carcinoma (cSCC) samples, a frequent non-melanoma skin cancer, by annotating 47,392 cell nuclei across 5  cell types in 21 WSIs and segmenting tumor regions in 144 WSIs. Histo-Miner employs convolutional neural networks and vision transformers for nucleus segmentation and classification, as well as tumor region segmentation. Performance of trained models positively compares to state of the art with multi-class Panoptic Quality (mPQ) of 0.569 for nucleus segmentation, macro-averaged F1 of 0.832 for nucleus classification and mean Intersection over Union (mIoU) of \deleted[id=lu]{0.884}\added[id=lu]{0.907} for tumor region segmentation. From these output, the pipeline can generate a compact feature vector summarizing tissue morphology and cellular interactions, which can be used for various downstream tasks. As an exemplary use-case, we deploy Histo-Miner to predict cSCC patient response to immunotherapy based on pre-treatment WSIs from 45 patients.  Histo-Miner predicts patient response with mean area under ROC curve of 0.755 $\pm$ 0.091 over cross-validation, and identifies percentages of lymphocytes, the granulocyte to lymphocyte ratio in tumor vicinity and the distances between granulocytes and plasma cells in tumors as predictive features for therapy response. This highlights the applicability of Histo-Miner to clinically relevant scenarios, providing direct interpretation of the classification and insights into the underlying biology. Importantly, Histo-Miner is designed to allow for its use on other cancer types and on other training datasets. Our tool and datasets are available through our github repository: \href{https://github.com/bozeklab/Histo-Miner}{https://github.com/bozeklab/histo-miner}. 
%/keywords{vision transformers \and convolutional neural networks \and nuclei segmentation \and nuclei classification \and tumor segmentation \and cutaneous squameous cell carcinoma \and immune checkpoint inhibitors}
\end{abstract}

\restoregeometry

\begin{figure}[]
    \includegraphics[width=1\textwidth]{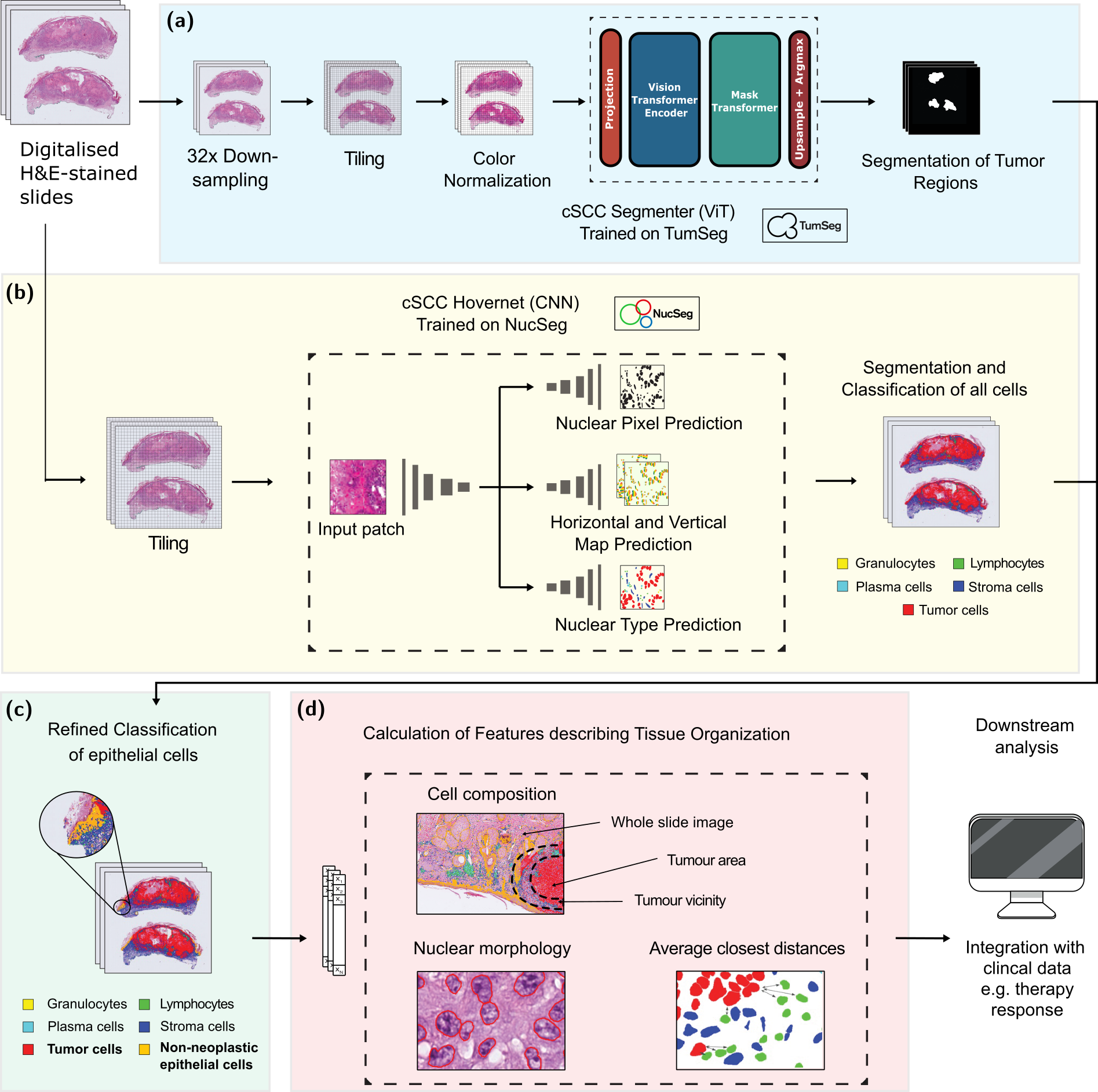}
    \caption[Caption Title]{ \normalsize \textbf{Overview of Histo-Miner pipeline.} 
    \small  The pipeline uses WSI from cSCC patient as input. \textbf{(a) \& (b)} \added[id=lu]{During inference, WSI images are tiled into patches and undergo pre-processing pipelines (see \textbf{Methods}). After pre-processing, SCC Segmenter performs tumor region binary segmentation on processed patches and SCC Hovernet segments and classifies cell nuclei.} \deleted[id=lu]{Images are processed via two different pre-processing pipelines that involve downsampling, tiling and color normalization. After the first pre-processing pipeline, resulting images are input to SCC Segmenter for tumor region binary segmentation. In parallel, images after the second pre-processing pipeline are input to SCC Hovernet to segment and classify cell nuclei.}  \textbf{(c)}  Using the output of SCC Segmenter, cell classification is refined by adding a new cell class: non-neoplastic epithelial. This last result is saved in a json text file. A visualization of resulting annotations is provided in  \textbf{Fig. \ref{fig2}}. \textbf{(d)}  Using refined segmentation and classification of nuclei, together with segmentation of tumor regions, we calculate features that describe the tissue organization. Example features include e.g. percentage of lymphocytes in the vicinity of the tumor and average closest distance between tumor and lymphocyte cells. A list of all 317 calculated features is provided in \textbf{Supplementary Data}.}
    \label{fig1}
\end{figure}

\section*{\added[id=lu]{Author Summary}}

\added[id=lu]{Digital pathology is transforming how we study disease by turning tissue samples into high-resolution images that capture the architecture of entire tumors. However, these images are vast and complex, making it difficult to extract meaningful clinical insights without advanced computational tools. In this work, we present Histo-Miner, a framework designed to systematically analyze these images at multiple levels of detail—from one single cell to entire tissue regions. We apply this approach to cutaneous squamous cell carcinoma, a common form of skin cancer, demonstrating how large-scale tissue data can be mined for biological insights. Our method identifies and characterizes different types of cells, maps how they are organized within tumor areas, and connects these patterns to patient outcomes. Through this lens, we uncover subtle features of the tissue environment that may influence how patients respond to therapy. We find that the most informative features describe the presence and balance of different types of immune system cells, and how these cells are spatially arranged within the tissue. Beyond its immediate findings, Histo-Miner, provides openly available data and tools that aim to make large-scale tissue analysis more interpretable, reproducible, and transferable to other diseases.}

\section*{Introduction}
Digital pathology slide scanners and advancements in computer vision allow for automation of diagnostic pathology tasks. Hematoxylin and Eosin (H\&E) staining \cite{Wittekind2003} is widely used in pathology and represents a standard that both the classical and digital pathology are based on. The resulting tissue scans are called Whole-Slide Images (WSIs). Given the large size of WSIs containing thousands to millions of cells, automated methods for WSI analysis are indispensable to systematically and comprehensively quantify their content. A large panel of tasks can be performed by machine learning and deep learning models on this type of images: segmentation of nuclei and tumors in the WSIs \cite{Rijthoven2021,Shui2024}, image classification \cite{Zheng2022}, or discovery of new biomarkers \cite{Nahhas2023}. Importantly, such methods automate time consuming intermediary tasks, such as cell counting, that allows the practitioner to focus on diagnosis and interpretation \cite{Pantanowitz2020,Viswanathan2022}.   \\

While there is a range of datasets and methods in digital pathology \cite{Cardoso2022,Lu2023,Chen2024,Saldanha2023}, few of them are dedicated to skin and \added[id=lu]{non-melanoma} skin tumors. Skin differs from other tissues in its unique structure, composition, and function, which presents specific challenges for digital pathology methods. The skin consists of multiple distinct layers, including the epidermis, dermis, and subcutaneous tissue, each with varying cell types, densities, and extracellular matrices. These variations lead to textural patterns and coloration that are unique to WSIs of skin. \deleted[id=lu]{Digital pathology algorithms developed for other tissues may not accurately segment and recognize morphological elements of skin}. Therefore, specialized methods tailored to the unique characteristics of skin tissue are necessary for reliable digital pathology in dermatology. \\ 

Here, we focus on cutaneous squamous cell carcinoma (cSCC) - the second most common form of \added[id=lu]{non-melanoma} skin cancer in the USA and widely spread worldwide \cite{Howell2023}. While the majority of cSCCs can be cured by surgery alone, 5-10\% of cSCC patients experience disease recurrence or metastases, requiring systemic treatments \cite{Winge2023}. While the effects of chemotherapy are very limited, systemic treatment with immune checkpoint inhibition (CPI) has emerged as a promising alternative. However, up to half of patients do not respond to the immunotherapy. To date, it is impossible to predict, which patients have a high chance  of response to CPI and which patients may require other/additional treatment modalities \cite{Winge2023}. Quantitative methods for analysis of cSCC patient samples would provide more insights into the morphological variability of this tumor type and could potentially allow for identification of morphological markers linked to the patient risk of progression. \\ 

We propose a deep learning-based pipeline, Histo-Miner with openly available code and datasets, for development and analysis of cSCC samples. We generated a dataset called NucSeg containing manually annotated class labels and segmentation masks of 47,392 cell nuclei from 21 WSIs of cSCC. Furthermore, we generated TumSeg dataset, containing binary segmentation masks of tumor regions in 144 WSIs of cSCC. Our pipeline performs segmentation and classification of cell types into 6 different classes (granulocytes, lymphocytes, plasma cells, stromal cells, tumor cells and epithelial cells) and tumor region segmentation, both using deep learning models trained on our datasets. Histo-Miner uses the segmentation and classification results to encode WSIs of cSCC into features describing tissue morphology, organization, and cellular interactions. The code is open-source, customizable, and each part (tumor segmentation, cell type identification) can be used separately to fit user needs. The training datasets, as well as the models weights used in the intermediary steps are publicly available (See \textbf{Data Availability} and \textbf{Code Availability} sections). \\ 

We finally tested our Histo-Miner pipeline to predict cSCC patient response to immunotherapy. Immunotherapy with anti-PD1 antibodies is the major treatment for patients with advanced cSCC, but currently no predictive biomarkers are established to identify patients with a high likelihood of therapy failure. We generated the CPI dataset including 45 skin WSIs of 45 patients, before they received immunotherapy treatment, and annotated these slides as responder or non responder to the treatment. Using the features produced by Histo-Miner we classified patient response and found interpretable features explaining model choices. These features provide insights into biological factors favoring treatment response. The CPI dataset and the feature list are publicly available (See \textbf{Data Availability} section). 

\section*{Methods}

\subsection*{Histo-Miner pipeline description}
To describe the tissue organization and composition of cSCC, and obtain detailed information on the histomorphology of these tumors we developed our pipeline, Histo-Miner (\textbf{Fig. \ref{fig1}}). It uses both cell nuclei and tumor segmentation as first steps to quantitatively describe tumor sample morphology. \\

In the pre-processing pipeline of SCC Segmenter, the images are  downsampled, tiled into patches and the patches are normalized using mean and standard deviation of RGB pixel values of ImageNet 1K (see our github repository for implementation details). SCC Hovernet is trained with data augmentation for model generalization \cite{Graham2019} and then input patches don't need color normalization. To capture cellular heterogeneity, the cell nuclei are segmented and classified into: granulocytes, lymphocytes, plasma cells, stromal cells, and tumor cells. Cell nuclei segmentation and classification is performed with Hovernet convolutional network  \cite{Graham2019} trained on a manually annotated set of cSCC WSIs  (NucSeg). This model shows better instance segmentation and improved Panoptic Quality (PQ) compared to other recent H\&E segmentation models  \cite{Graham2019,Gamper2020}. The model is open source and allows users to train and adapt it.  Our pipeline additionally includes Segmenter vision transformer network to segment tumor regions. This vision transformer model outperforms other models in several benchmark tasks of instance and semantic segmentation \cite{Strudel2021,Zhou2019,Mottaghi2014}. We trained the Segmenter model on our TumSeg dataset to perform a binary pixel-wise classification tumor and non-tumor regions of the same WSIs that were used for Hovernet inference. We name the resulting trained networks SCC Hovernet and SCC Segmenter.  \\ 

\added[id=lu]{In the pre-processing pipeline of SCC Segmenter, the images are first downsampled, then tiled into patches and the patches are normalized using mean and standard deviation of RGB pixel values of ImageNet 1K (see our github repository for implementation details). In the pre-processing pipeline of SCC Hovernet, WSI input are only tiled into patches. In both cases,  only patches with tissue are kept for prediction (at least one pixel of tissue) after tiling.} \\

\begin{figure}[]
    \includegraphics[width=0.96\textwidth]{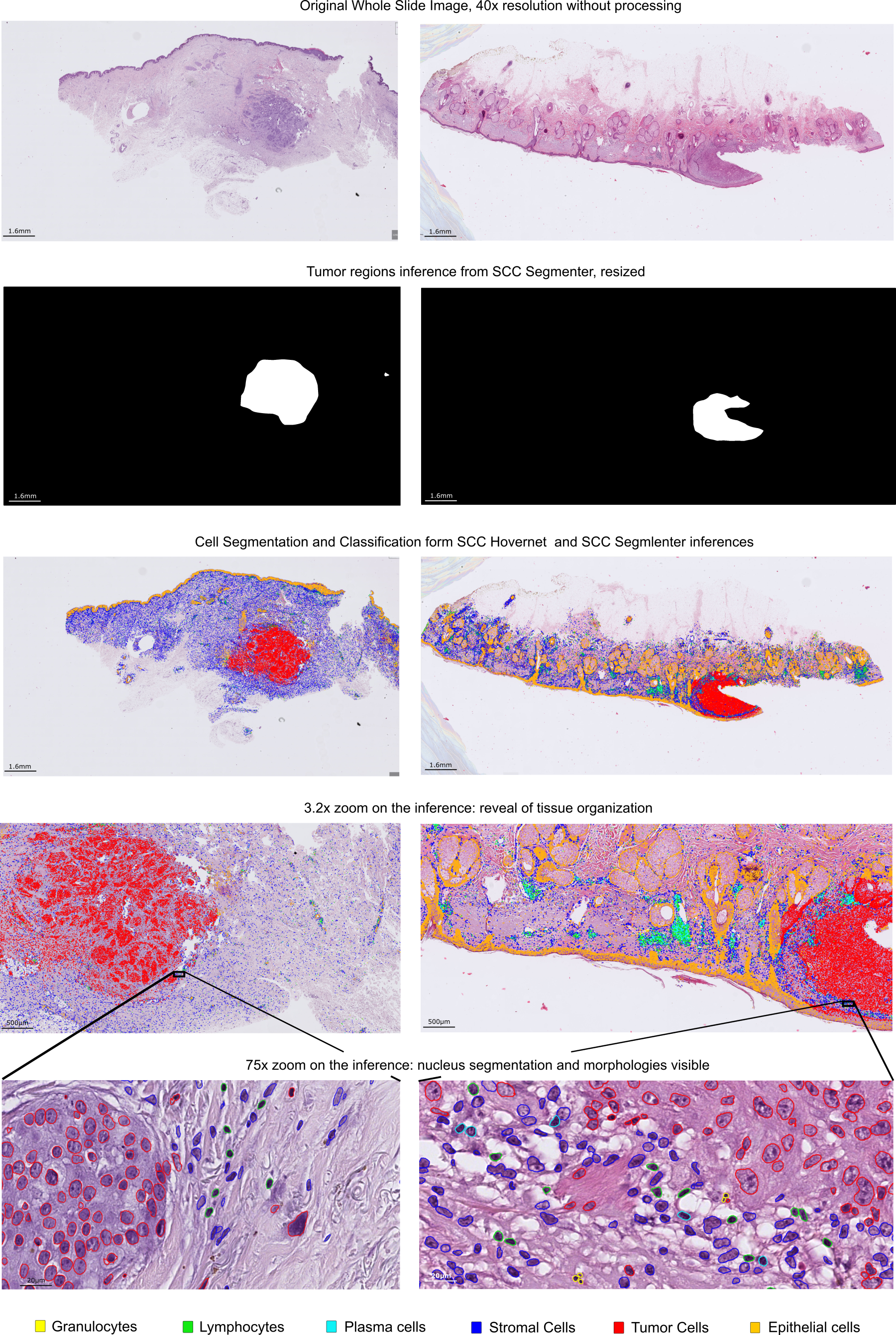}
    \caption[Caption Title]{ \normalsize \textbf{Predictions of SCC Hovernet and SCC Segmenter models.}
    \small  The different images correspond to different steps of the Histo-Miner pipeline as depicted in \textbf{Fig. \ref{fig1}}. 2 WSIs of 2 different patients from 2 different cohorts are shown. The H\&E staining differs between the slides showing varying hues of blue and pink. After predicting the tumor area with SCC Segmenter,  Histo-Miner segments and then classifies cells into five different classes: granulocytes, lymphocytes, plasma cells, stromal cells, tumor cells. Using tumor segmentation, tumor cells detected outside tumor regions are re-classified as non-neoplastic. The cell nuclei segmentation and classification illustrate sample organization at tissue level (3x zoom), or at cell level (75x zoom). Based on segmentation results Histo-Miner calculates features describing cell-level and tissue-level tumor organization. Also in the case of damaged sample (one part of the tumor is missing in the WSI on the right), the model is not hallucinating segmentation of the remaining parts of the sample.} 
    \label{fig2}
\end{figure}

After determining tumor areas using SCC Segmenter, the results of the SCC Hovernet cell nuclei classification are updated to add a new cell class as follows: all the nuclei predicted as tumor cells outside of the predicted tumor regions are reclassified as healthy epithelial. The reason for this update is that healthy epithelial cells and tumor cells have similar morphologies and are impossible to discriminate without a broader context and information about the tissue structure (see \textbf{Supplementary Fig.1}). Example visualization of the inference of the two methods is shown in \textbf{Fig. \ref{fig2}}.  \\

%\afterpage{\clearpage} 

Results of segmentation are input to tissue analysis part of our pipeline. In this part  we perform calculation of 317 features that describe and encode the tissue samples. Example features include percentages of cells of specific class anywhere in the sample as well as inside the tumor regions. For every pair of cell classes X and Y, we also calculate the average distance of the closest cell of class Y to a cell of class X inside the tumor regions. This feature describes the topology of the tissue and the interactions between cell classes. \\

An exhaustive list of the calculated features is available in \textbf{Supplementary Data}. We do not consider absolute numbers of cells of a given class as a feature, but cell densities and percentages, as this metric is dependent on the WSI size and not the structure of the tissue itself. All the features are stored in a light json file, which results in encoding and compressing a WSI of multiple GB into a text file of 3.7 KB. These features are a convenient WSI representation for any downstream analysis. All the different steps of the pipeline can be run separately as well as configured to fit specific needs. An example of use case,  predicting response of cSCC patients to immunotherapy, is provided in the following sections.

\subsection*{NucSeg and TumSeg datasets descriptions}
\begin{figure}[]
    \centering
    \includegraphics[width=0.95\textwidth]{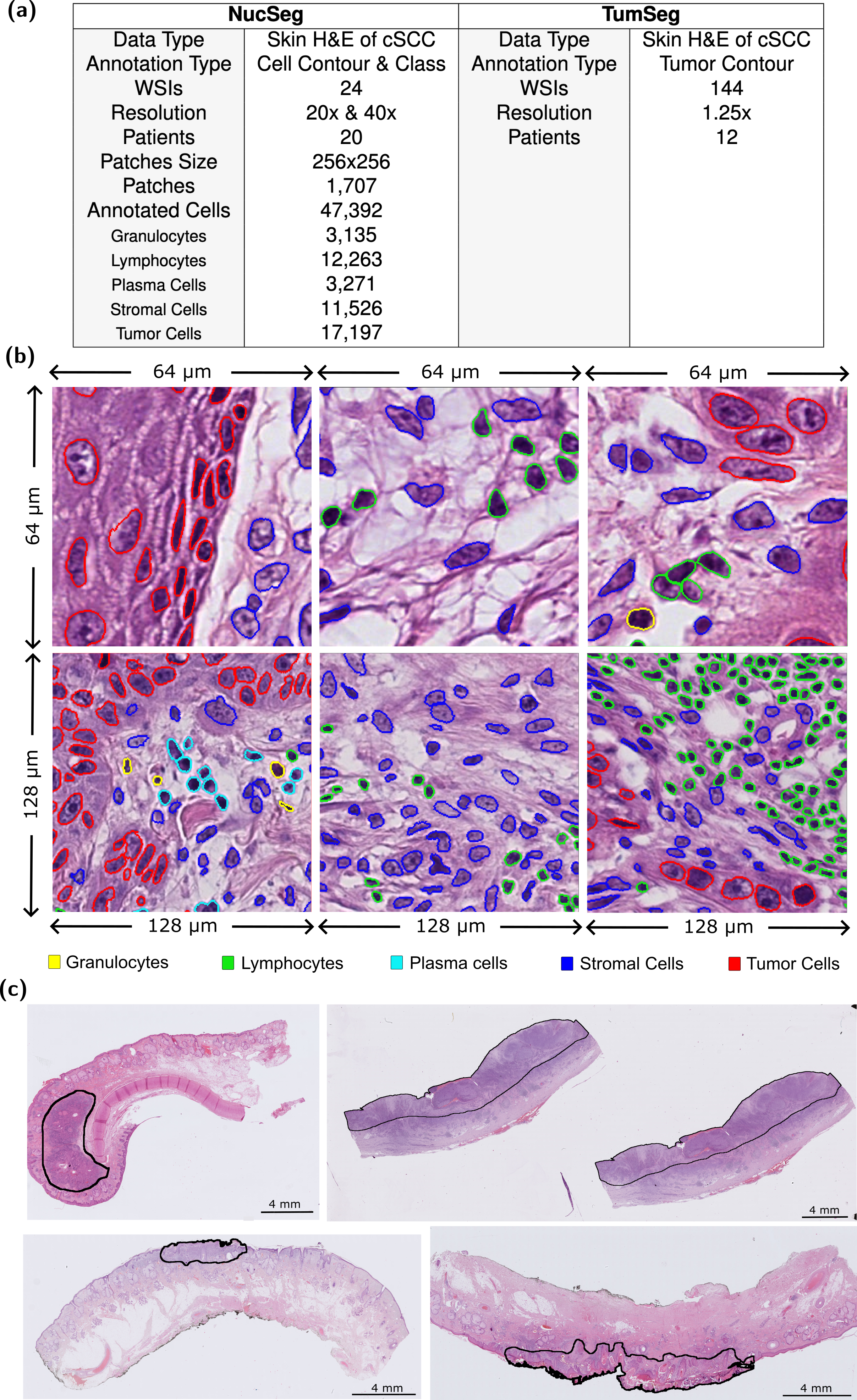}
    \caption[Caption Title]{ \normalsize \textbf{Visualization of samples from NucSeg and TumSeg training datasets.}
    \small \textbf{(a)} Overview of both datasets \textbf{(b)} Visualization of NucSeg training dataset. 47,392 cell nuclei from 1,707 H\&E non-overlapping patches were segmented and classified. \textbf{(c)} Visualization of TumSeg training dataset. Tumor region are segmented by 2 experts on 144 WSIs.}
    %\vspace{-1cm}
    \label{fig3}
\end{figure}

To enable segmentation and cell nucleus type classification for cSCC, we assembled 21 WSIs of H\&E-stained tissue sections of 20 cSCC patients from the University Hospital Cologne. The images were acquired using a NanoZoomer Slide Scanner (Hamamatsu). In the images the nuclei contours were marked and assigned to five cell types: granulocytes, lymphocytes, plasma cells, stromal cells, and tumor cells. 1,707 H\&E non-overlapping patches of 256x256 pixels, with 40x and 20x \replaced[id=lu]{resolutions}{magnifications}, have been manually annotated by two pathology experts. \added[id=lu]{To ensure annotation consistency across the distributed workflow, ambiguous morphological patterns were subjected to joint consensus review and compared to IHC staining on validation slides.} 47,392 nuclei were labeled (classified and segmented) in total (3,135 granulocytes, 12,263 lymphocytes, 3,271 plasma cells, 11,526 stromal cells, 17,197 tumor cells), see \textbf{Fig. \ref{fig3}a} and \textbf{Fig. \ref{fig3}b}. \\

The annotations consist of two groundtruth patches for each H\&E patch. The first annotation is an instance segmentation of each nucleus. A unique value is attributed to pixels belonging to the same nucleus instance, where every nucleus is assigned a different value. Annotation of nuclei instances allows to differentiate touching instances based on their instance pixel values. All pixels outside nucleus (background) are given the pixel value 0. The second annotation is a type map of the nucleus, in which the pixels belonging to nucleus of the same class have the same pixel values. The dataset is also available as 6,816 patches of  560x560 pixels with 70\% overlap in a 5D numpy array according to the Hovernet data format requirements. \\  

To build a tumor segmenter algorithm, we additionally assembled 144 WSIs of 125 cSCC patients from 3 medical centers - Bonn, Cologne and Munich. These WSIs are a subset of the dataset described in \cite{Pisula2024_2}. All tumor regions were labeled by two pathology experts. The WSIs were originally at \replaced[id=lu]{resolution}{magnification} 40x and downsampled 32 times (reaching a final \replaced[id=lu]{resolution}{magnification} of 1.25x) to enable to fit them into memory during model training (\textbf{Fig. \ref{fig3}a}). \added[id=lu]{The resulting image–tumor annotation pairs constitute our TumSeg dataset, more precisely, downsampled WSIs and the binary segmentation masks of their tumor regions.  We also assembled and labeled 32 slides from 25 other patients to create a test set for model evaluation (see \textbf{Results} section).} Examples of WSI tumor segmentation annotations are displayed in \textbf{Fig. \ref{fig3}c}. \deleted[id=lu]{WSIs and the binary segmentation masks of their tumor regions are part of TumSeg dataset}. In addition, anonymized patient IDs and medical centers IDs are available as metadata. \\  

These 2  datasets, NucSeg and TumSeg, are openly available in our Zenodo repository  \href{https://zenodo.org/records/8362593?preview=1&token=eyJhbGciOiJIUzUxMiJ9.eyJpZCI6ImMyNWMwNTU2LThlMDQtNDkwNS1hN2ZkLWMzYjAyNzg4ZmJlZSIsImRhdGEiOnt9LCJyYW5kb20iOiJiZjQzZThlOGJjMzEwMWJjZjk2YzNiZTY4MGI0MDhlNSJ9.NEULZTiXnprYpvbzFrQLQyKdF7ocHaXwkwup8LyIMefQB5ydyG-BasrZ9fkbfXmaoPNS0l01BrZMcwaH5DO6oA}{link}.

%\afterpage{\clearpage} 

\subsection*{Histo-Miner deep learning models}
Histo-Miner implementation includes deep learning models trained with our custom datasets. Histo-Miner users can utilize the weights of these models to perform similar inferences on their own datasets, re-train these models through Histo-Miner implementation directly and edit model architectures and hyperparameters for further development. To fit user needs, it is possible to use only specific blocks of the pipeline - such as inference of the deep learning models - instead of using the whole process until calculation of tissue features. \\

We performed segmentation and classification of segmented cell nuclei using Hovernet model \cite{Graham2019}, which we selected based on its performance and ease of use.  The model is a convolutional neural network containing encoder and decoder parts. The semantic segmentation of tumor region on the WSIs was achieved using Segmenter, a vision transformer model \cite{Strudel2021}.  Segmenter  is a collection of architectures with varying size, composed of encoder and decoder parts. In Histo-Miner pipeline we use the Seg-L-Mask/16 segmenter variant, achieving better results than the base model but requiring more GPU memory.   \\

\subsection*{Training SCC Hovernet}
To segment and classify cell nuclei into different cell classes (granulocytes, lymphocytes, plasma cells, stromal cells, and tumor cells) we trained Hovernet network with our dataset NucSeg. The training was performed on 2 80GB A100 NVIDIA GPUs (Ampere micro-architecture). We used Hovernet with the encoder pretrained on ImageNet 21k. A second pre-training of 150 epochs was performed on a not-curated dataset of H\&E nucleus segmented and classified, also made openly available. This dataset resemble NucSeg, consists of the same classes, but the segmentation and classification were mostly automatized and not fully corrected by human experts. During the first 50 epochs, the encoder weights were frozen and during the following 100  epochs all weights were updated. Finally, the main training of 250 epochs (first 50 epochs with frozen encoder weights followed by 200 epochs with all weights updated) was performed on NucSeg dataset. We used Adam optimization algorithm \cite{Kingma2014} with initial learning rate $\gamma_{0}$ of $10^{-4}$ which was reduced to $10^{-5}$ after 25 epochs. We used the same loss functions as in the original Hovernet model. \added[id=lu]{The global loss function is an addition of three losses for each of the three branches of the Hovernet Model. These branches account, respectively, for the nuclear pixel classification task, the binary segmentation of nucleus, and the horizontal and vertical distance map used to separate touching instances \cite{Graham2019}.} \replaced[id=lu]{We describe the loss function in detail in}{We list loss functions in} \textbf{Supplementary Eq.1 - 4}. We used the following data augmentations during training: image flips, rotations, Gaussian blurs and median blurs according to the original Hovernet implementation. \\

We trained NucSeg dataset including 5,968 patches of size 540x540 pixels with 70\% overlap. Our validation set contained 848 patches of size 540x540 pixels with 70\% overlap. We kept the same set of hyperparameters as the original implementation \cite{Graham2019}, except that we doubled the training batch size for our last training step of 200 epochs (see above). We tested different training strategies combining network pre-trained on ImageNet 21k, pre-trained on the not-curated H\&E nucleus dataset, and trained from scratch. Performance of resulting models was highest when the model was first pre-trained on ImageNet 21k, then pre-trained on the not-curated H\&E nucleus dataset before fine-tuning. The choice of final model was based on maximizing the panoptic segmentation task performance.

\subsection*{Training SCC Segmenter}
We trained Segmenter model with our dataset TumSeg. The training was performed on 2 80GB A100 NVIDIA GPUs (same as for Hovernet training). We used Vision Transformer pre-trained on  ImageNet 21k \cite{Steiner2021} and fine-tuned it on our dataset TumSeg. We followed the data augmentation pipeline from the semantic segmentation library MMSegmentation \cite{mmseg2020}.  It consists of random resizing of the image to a ratio between 0.5 and 2.0, random left-right flipping, and normalization of the images based on mean and standard deviation of pixel values of ImageNet 1k. We tested other normalization strategies, e.g taking mean and standard deviation of training dataset which resulted in worse model performance.  We trained for 1,786 epochs (50,000 iterations) using Stochastic Gradient Descent as optimization method with learning rate following a polynomial decay scheme. Considering $\gamma$ the learning rate at the current iteration number, and $\gamma_{0}$ the base learning rate, the decay is defined as  $\gamma = \gamma_{0} ({\frac{1 - N_{iter}}{N_{total}})^{0.9}}$ where $N_{iter}$ and $N_{total}$ represent the current iteration number and the total iteration number, respectively. We set $\gamma_{0}$ to $10^{-3}$.  We used cross-entropy without weight re-balancing as the loss function. \\

The model was trained on randomly chosen 115 slides of TumSeg dataset and the validation set contained the remaining 29 slides of the dataset. We performed hyperparameter grid-search to find the best set of hyperparameters as described in \textbf{Supplementary Table. 1}. The accuracy estimation was performed on the validation set due to the limited number of slides and lack of an independent test set.   \\

\subsection*{Tissue Analyser}
Within Histo-Miner, SCC Segmenter model segments tumor regions. SCC Hovernet model performs instance segmentation of cell nuclei and classifies them into different cell classes (granulocytes, lymphocytes, plasma cells, stromal cells, and tumor cells). Using both models’ predictions, we refine the cell classification and add one more class among the possible predictions, the healthy epithelial class. In fact, healthy epithelial cells and tumor cells are hard to discriminate without a broader context and information about the tissue structure. Healthy epithelial cells and tumor cells have similar morphologies. To distinguish these two cell types, we added one refinement step: all the nuclei predicted as belonging to tumor cells by SCC Hovernet, located outside the tumor regions predicted by Segmenter, are reclassified as epithelial. The result of the classification update is visible in \textbf{Fig. \ref{fig2}}.   \\

The updated cell nuclei segmentation and classification as well as the tissue segmentation are input to our Tissue Analyser part of the pipeline. Here we calculate 317 features capturing various aspects of tissue morphology and spatial organization. The features (see \textbf{Supplementary Data}) include: percentages of given cell types, composition of the tumor margin, ratio between cell types, repartition of a given cell type outside, inside and within the tumor margin. The final feature vector is a light but information-rich representation of the cSCC WSI for further downstream analyses.  \\

One of our features is the average distance of the closest cell of a given class X - source class - to the closest cell of a given class Y - target class - inside the tumor regions. The average distance is calculated as shown in \textbf{Eq. \ref{eqdistance}} and through the following steps: \textbf{1-} We generate a rectangle that defines the initial search area. The rectangle height is 5\% of the tumor bounding box height and width 5\% of the tumor bounding box width. It is centered around the nucleus of the source cell class. \textbf{2-} We verify if there is at least one nucleus of the target cell class inside this search area. \textbf{3-} If there is at least one, we calculate all distances between source and target cells and keep the smallest one. If there is no nucleus of target cell class we increase the search area until we find at least one nucleus of target cell class in the search area. The search area cannot extend to other tumor areas. \textbf{4-} We perform steps 1-3 for all nuclei of the source class. A quantitative explanation is available in \textbf{Algorithm. \ref{pseudocode-distance}} (all the memory optimization steps are skipped for readability). The increase of search area for each iteration was optimized to reduce calculation time. For computation optimization reasons, the search area is a rectangle and not a circle. Indeed, one of the main reason is that searching for coordinates in a rectangle is faster than searching for coordinates in a circle  (only comparisons instead of subtractions and multiplications). In some specific cases, using a rectangle search area can lead to overestimation of the distance. Description of these cases, probability of overestimation, and bounding of overestimation are described in \textbf{Supplementary Fig. 3}. This probability of overestimation decreases drastically with the number of cells in the tumor. For instance, we can calculate that for a squared search area of side length $2r_{1}$ and origin $0$, if $N$ cells are in the circle of radius $\sqrt{2} r_{1}$ and origin $0$ (the square is inscribed in this circle), the probability of overestimation is $P(error_{N})_{2} = 0.022$ for $N=2$ and $P(error_{N})_{10} = 2.8 \mathrm{x} 10^{-4}$ for $N=10$. These distances describe the interactions between different cell types inside the tissue. They are calculated for granulocytes, lymphocytes, plasma cells, and tumor cells to assess the organization of the tissue regarding the intensity of the immune response. 

% Finally the  overestimation impact is limited on the mean of all the closest distances. Additionally, the aspect ratio of tumor bounding box is limited by biological constraints (a tumor cannot be a layer of one cell wide), reducing the error as the search area is based on the tumor bounding box, see \textbf{Algorithm.\ref{pseudocode-distance}}. 
%  

% Could add it here as well
% $ \sqrt{lenght^{2} + width^{2}} - width $ , so up to $ 41\%$ of the lenght of the search area if the search area is a square
\vspace{0.5cm}

\Needspace{20\baselineskip}
% \begin{samepage}
    
% \begin{minipage}{\textwidth}

\begin{equation}
\bar d_{closest_{{c_{A}} , c_{B}}}= \frac{1}{n_{c_{A}}} \sum_{\substack{(x_{i}, y_{i}) \in E_{c_{A}}}} \min\left(\sqrt{(x_i-x_k)^2+(y_i-y_k)^2}
\right), 
\label{eqdistance}
\end{equation}
%\vspace{0.02cm}
\[
 \forall (x_{k}, y_{k}) \in \Bigl( E_{c_{B}} \cap E_{(x, y) \in \mathcal{N}^{2}  \mid \  
\left\{
\scalebox{0.8}{
    \(
    \begin{array}{l}
    \phantom{}0.05 \ \lambda_{0} l_{t} - x_{i} \leq x \leq 0.05 \ \lambda_{0} l_{t} + x_{i} \\   
    \phantom{}0.05 \ \lambda_  {0} w_{t} - y_{i} \leq y \leq 0.05 \ \lambda_{0} w_{t} + y_{i} 
    \end{array}
    \)
}
\right.}
^{ \lambda_{0}} 
\Bigr)
\]
%\vspace{0.02cm}
\[
\text{with} \ \lambda_{0} \ \textrm{verifying:}
\]
\[
\forall \lambda \in \mathcal{N}^{*} \mid \Bigl( E_{c_{B}} \cap E_{(x, y) \in \mathcal{N}^{2} \ \mid \ 
\left\{
\scalebox{0.8}{
    \(
    \begin{array}{l} 
    \phantom{}0.05 \ \lambda l_{t} - x_{i} \leq x \leq 0.05 \ \lambda l_{t} + x_{i} \\   
    \phantom{}0.05 \ \lambda w_{t} - y_{i} \leq y \leq 0.05 \ \lambda w_{t} + y_{i} \\ 
    \end{array}
    \)
} 
\right.}
^{ \lambda}  
\Bigr)
\ne \{  \emptyset  \}
, \ \lambda_{0} = \min \left(\lambda \right) 
\]
\vspace{0.2cm} 

\begin{adjustwidth}{30pt}{30pt}
\noindent  \small where $\lambda \in \mathcal{N}^{*}$ coefficient of increase of sides length of search rectangle, $l_{t}$ and $w_{t}$ length and width of the bounding box of the tumor region, $E_{c_{a}}$ set of coordinates in $\mathcal{N}^{2}$ of all cells centroid of class A (source class), $E_{c_{b}}$ set of coordinates in $\mathcal{N}^{2}$ of all cells centroid of class B (target class), $n_{c_{A}}$ number of cells of class A. \\  \\
$E^{ \lambda}$ is the set of points inside the search rectangle of coefficient $\lambda$ and $E^{ \lambda_{0}}$ is the set of points inside the smallest search rectangle that contains at least one centroid of cell of class B. 
\end{adjustwidth}

% \end{samepage}
% \end{minipage}

\begin{algorithm}[]
    \caption{Minimum Distance Calculation Pseudo-Code}
    \fontsize{9}{10}
    \textbf{Require:} $classjson$: file with centroid and type of all WSI’s cells \\
    \textbf{Require:} $maskmap$: segmentation of WSI’s tumor regions  \\
    \textbf{Require:} $selected\_classes$: classes for which we want to calculate distances, $f$: 
    acceleration of increase parameter - here $f = 0.05$  \\
    Start $C(n, 2) =  \frac{n(n-1)}{2}$ parallel processes, $n = \mathbf{len}(selected\_classes)$ \\
    In each process a different pair source class / target class is defined \\
    \textbf{For each} process: \\
    \phantom{tab} $allmindist$ = \textbf{list()} \\
    \phantom{tab} \textbf{For all} cells of source class: \\
    \phantom{tabtab} $dist\_list$ = \textbf{list()} \\ 
    \phantom{tabtab} Check tumor ID of source cell \\
    \phantom{tabtab} List all target class cells $centroid$ $coordinates$ in the same Tumor \\ 
    \phantom{tabtab} Calculate length $l_{t}$ and width $w_{t}$ of the bounding box of the Tumor \\
    \phantom{tabtab} $\lambda = 1$ \\ 
    \phantom{tabtab} $kept\_targets$ = \textbf{list()} \\ 
    \phantom{tabtab} \textbf{While} $kept\_targets$ cells list is \textbf{empty}: \\ 
    \phantom{tabtabtab} Construct search zone around source cell with $length = f \lambda  l_{t}$ and $width = f \lambda  w_{t}$ \\ 
    \phantom{tabtabtab} Append to $kept\_targets$  \textbf{all} target cells $centroid$ $coordinates$ inside the search area \\
    \phantom{tabtabtab} $\lambda=\lambda+1$ \\
    \phantom{tabtab} \textbf{For all} cell in $kept\_targets$: \\
    \phantom{tabtabtab} $dist = \sqrt{(x_i-x_k)^2+(y_i-y_k)^2}$ with $(x_{i}, y_{i})$ coordinates of cells of source class and \\
    \phantom{tabtabtab} $(x_{k}, y_{k})$ coordinates of cells of target class\\
    \phantom{tabtabtab} $dist\_list \mathbf{.append} (dist)$ \\
    \phantom{tabtab} $min\_dist$ = \textbf{min}($dist\_list$) \\
    \phantom{tabtzb} $allmindist$\textbf{.append}($min\_dist$) \\
    \phantom{tab}  Put item $avgdist = \textbf{sum}(allmindist) / \textbf{len}(allmindist)$ in process queue \\
    Ouput a list of all $avgdist$ gathered items
    \label{pseudocode-distance}
\end{algorithm}

\vspace{0.5cm} 
Other features include ratios of cell types and percentages of cells of a given cell type in the vicinity of the tumor, in the tumor regions or in the whole WSI.  The vicinity of tumor is defined as a  $1 \textrm{mm}$-wide area around the tumor \cite{Hendry2017}. In the case of ratios, to limit outlayers we define ratio $\eta = \frac{\log{(n_{c_{A}})} + \epsilon}{\log{(n_{c_{B}})} + \epsilon}$. with $n_{c_{A}}$  number of cells of class A, $n_{c_{B}}$ number of cells of class B and $\epsilon = 10^{-3}$, smoothness parameter. The full list of features is in the  \textbf{Supplementary Data}.

\subsection*{Feature selection to predict response of cSCC patients to immunotherapy}

We collected WSIs from 45 patients (one per patient) with cSCC skin cancer, from 6 medical centers - Bonn, Cologne, Dortmund, Munich, Oberhausen and Salzburg \added[id=jo]{taken} before \deleted[id=jo]{they received}\added[id=jo]{administration of} immune checkpoint inhibitors treatment. 28 of them were classified as responders, showing partial response (PR) or complete response (CR), i.e. tumor shrinkage. 17 patients were classified as non-responders, showing stable disease (SD) or progressive disease (PD) states. More specifically, CR means disappearance of all lesions; PR: 50\% or more in decrease of total tumor size; SD: $<$50\% decrease and/or $<$25\% increase of one/several tumor lesions; PD: $>$25\% increase of one/several tumor lesions or new lesions. The classification was determined by a dedicated review of the clinical and radiological imaging by at least two observations not less than 4 weeks apart (following World Health Organization handbook for reporting results of cancer treatment). This collection of classified WSIs represents our fourth dataset, called CPI (see \textbf{Data availability} section). \\  

We used Histo-Miner to extract tissue representative features from the WSIs of CPI dataset. Then, we trained and evaluated an XGBoost classifier \cite{Peng2005} for the task of classifying patients in their two categories based on the feature vectors. We evaluated XGBoost classifier through 3 fold cross-validation \added[id=lu]{of 2 splits, train and test (train containing 2/3rd of the data),} and performed feature selection \added[id=lu]{on the training split} within the cross-validation runs. Due to the limited number of samples in the dataset (45) we could not perform hyperparameter search within nested cross-validation so we kept the default set of hyperparameters for all the cross-validation runs.  Similarly, the low number of folds is constrained by the low number of samples. In fact, having too small validation folds would increase variance in evaluation. We used minimum Redundancy - Maximum Relevance (mRMR) feature selection method \cite{Chen2016} which is designed to find the smallest relevant subset of features (maximum relevance) while preventing highly correlated features to be part of this subset (minimum redundancy). All the 107 features kept for analysis are describing tissue structure, no nucleus morphology features (area, circularity) were included (See \textbf{Tissue Analyser} section in \textbf{Methods} for an in-depth description of the features). \\ 

To know how many features to keep we calculated the average balanced accuracy for each training keeping $ N \in [1, 107]$ features. We kept $N_{\text{best}} = 19$ features which corresponded to the best mean balanced accuracy across all runs.  Even if the number of selected features $N_{best}$ is the same for each run in the case of mRMR, the selection of features can vary.  Following \textbf{Eq. \ref{keepbestfeat}}, we identify the most representative features by selecting those with the highest occurrence counts ${c_{f_{k}}}$  across the selected feature sets in each cross-validation run. In the event of ties, we favor features with higher rankings in their respective sets.  To do so we first record the position of the feature in its set, its rank, and calculate its pre-score as $10^{N_{best} - rank}$, so a feature ranked first would have the highest pre-score. Then we add all the pre-scores for each cross-validation groups for a given feature and take the log of this sum to obtain the final score $s_{f_{k}}$.     

\Needspace{10\baselineskip}

\begin{equation}
A = \operatorname{concat} \{  fv_{split_{1}} , fv_{split_{2}} , ... , fv_{split_{L}}  \} 
\label{keepbestfeat}
\end{equation}
\[
c_{f_{k}} = \sum_{l=1}^{L} \sum_{n=1}^{N_{b}} \delta(A_{l,n}, f_{k})
\]
\[
s_{{f_{k}}} = \log (\sum_{l=1}^{L} 10^{{N_{best} - rank(f_{k})_{split_{l}}} })
\]

\vspace{0.2cm} 
\begin{adjustwidth}{30pt}{30pt}
\noindent  \small where  $fv_{split_{l}}$ = $[f_1,f_2,...f_{N_{b}}]_{split_{l}}$ the vector of $N_{best}$ selected features from split $l \in [1, L]$ of the cross-validation,  $rank(f_{k})_{split_{l}}$ the rank of feature $f_{k}$ selected from split $l \in [1, L]$   and  $\delta \text{ the Kronecker delta}$. 
\end{adjustwidth}
\vspace{0.5cm} 

Notably, mRMR prevents redundancy in feature selection, but aggregation of features from all cross-validation runs can include highly correlated features. \\

\deleted[id=lu]{Other feature selection methods were tested but led to a worse performing feature set. We tested Boruta \cite{Kursa2010} method which, unlike mRMR that finds the smallest relevant subset of features, aims at finding all features carrying information usable for prediction. With Boruta the number of selected features varied for each fold. Classification of CPI WSIs using Boruta selected feature had a balanced accuracy of $\bar {ba}_{N_{{BORUTA}}} = \bar {ba}_{[5,0,1]} = 0.711 \pm  0.011 $ across the 3 folds. For one of the fold Boruta was enable to select features, and this for any setting of maximum depth of the random tree. In the best scenario presented here the maximum depth was set to 1, the random state was set to 0  and all other hyperparameters were kept as default values. We also ranked features following Mann-Whitney U test scores \cite{Wilcoxon1945}. We then trained 107 XGBoost classifiers on the same training split, iteratively removing the least predictive feature each time, thus going from all 107 features down to the single most predictive feature (as we did for mRMR). The best balanced accuracy was in this case $\bar {ba}_{N_{MannWhitney}} = 0.715 \pm  0.069 $ with $N_{MannWhitney} = 6$.}

\section*{Results}

\subsection*{Nuclei segmentation and classification evaluation}
Accurate segmentation and classification of nuclei is necessary, to ensure the quality of downstream analyses involving features based on cell nuclei. We used Panoptic Quality metric \cite{Graham2019} to evaluate segmentation, as it has been shown that it is better suited for evaluation of instance segmentation than DICE2 (aggregation of DICE score for each instance) \cite{Vu2018} or aggregated Jaccard Index (AJI) metrics \cite{Kumar2017}. Indeed DICE2 is oversensitive to small changes in the prediction and AJI is over-sensitive to  failed detections as shown in \cite{Graham2019}. Following its definition, Panoptic Quality also assess detection performance.  \\   

We compare our cell nuclei segmentation and classification model to CellViT, current state of art model for segmentation  of cells in H\&E-stained WSIs \cite{Horst2024}. CellViT was originally trained on Pannuke dataset \cite{Gamper2020} a commonly used and recognized dataset for panoptic segmentation on H\&E stained tissues. Pannuke contains diverse types of cancer but only few skin cancer images, and without distinction of the type of skin cancer. Additionally, the dataset does not include granulocytes and plasma cells and the Pannuke-pretrained CellViT is not able to detect them. \\

\begin{figure}[]
    \centering
    \includegraphics[width=1\textwidth]{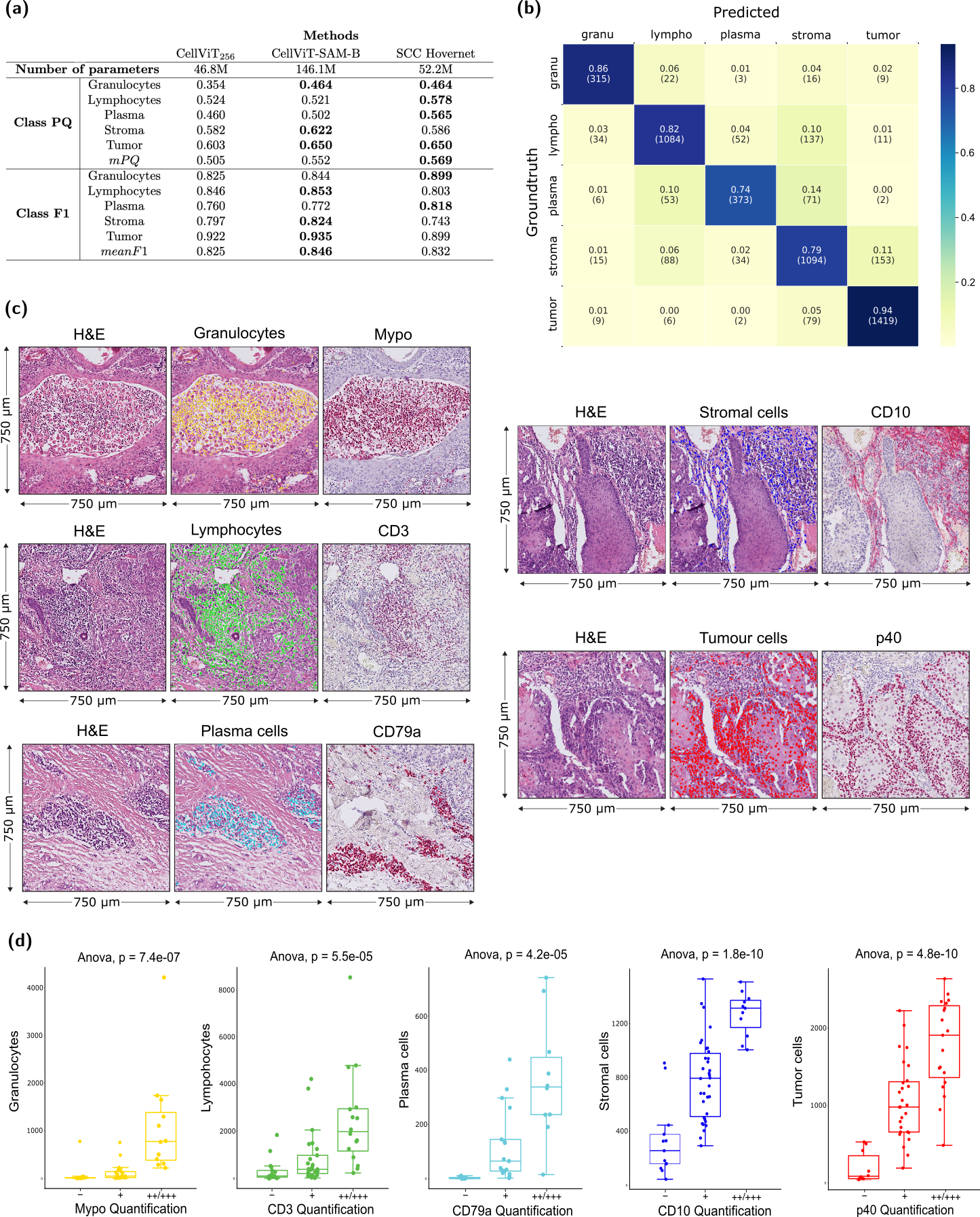}
    \caption[Caption Title]{ \normalsize \textbf{Validation of SCC Hovernet.}
    \small \textbf{(a)} Panoptic Quality  for each cell class of SCC Hovernet, light $\text{CellViT}_{256}$  and heavy CellViT-SAM-B, all trained on NucSeg. mPQ is the average of PQ for all classes. SCC Hovernet outperforms $\text{CellViT}_{256}$ for all classes and outperforms CellViT-SAM-B general mPQ performance. CellViT-SAM-B outperforms SCC Hovernet on general classification performance.  Taking into account segmentation, detection, classification tasks and weight of the models, SCC Hovernet is the preferred option.  \textbf{(b)} Confusion matrix from SCC Hovernet prediction on the validation set. The most representative class, tumor cells, is accurately classified 94\% of the time. The worst classification accuracy is of plasma cells: 74\%. \textbf{(c)} Examples of validation via immunohistochemistry showing H\&E and cell-type predictions (left and middle column) and staining for cell type markers of the same are in a consecutive section (right column) Mypo=Myeloperoxidase.  \textbf{(d)} Comparison of \added[id=lu]{manual} cell type \added[id=lu]{quantification}\deleted[id=lu]{counts} in immunohistochemistry slides (x-axis\added[id=lu]{; Mypo = Myeloperoxidase}) and computationally predicted cells (y-axis) in H\&E slides.}
    \label{fig4}
\end{figure}

We trained both models on NucSeg dataset including 5,968 patches of size 540x540 pixels with 70\% overlap. Our validation set contained 848 patches of size 540x540 pixels with 70\% overlap (see \textbf{Method} section for more details on training dataset and training procedure). No tiles were overlapping between the train and validation sets. We compared segmentation maps from SCC Hovernet and CellViT models applied to the same validation set in  \textbf{Fig. \ref{fig4}a}. The segmentation and classification results of each cell class are compared one by one.  \\

SCC Hovernet outperforms $\text{CellViT}_{256}$ in the detection and segmentation tasks for all cell classes. It also performs better or same on all cells nuclei classes except stromal cells, when compared to a larger CellViT model, CellViT-SAM-B trained on NucSeg. The average Panoptic Quality for all classes, mPQ of the SCC Hovernet is also higher compared to the CellViT-SAM-B trained on NucSeg. CellViT-SAM-B was too large to be trained on our 80GB A100 NVIDIA GPU directly so we used mixed-precision during training to fit it on one GPU (as no option to split the model on different GPUs was implemented). Mixed-precision has only limited impacts on models performance has shown in \cite{Dorrich2023}.  \\  

We also compared the 3 models on the classification task, considering only detected and paired cells (prediction and groundtruth). We evaluate F1 score for each class  (\textbf{Fig. \ref{fig4}a}). CellViT-SAM-B is the best classifier for 3 classes and SCC Hovernet for the 2 others, average of F1 over all classes is 0.825 for $\text{CellViT}_{256}$,  0.846 for CellViT-SAM-B and 0.832 for SCC Hovernet. A confusion matrix of SCC Hovernet classification on the validation set is shown in \textbf{Fig. \ref{fig4}b}. \\

Overall, SCC Hovernet slightly outperforms CellViT-SAM-B in segmentation and detection tasks but slightly under-performs CellViT-SAM-B in solely classification task. Nevertheless SCC Hovernet has 3 times less parameters \cite{Zhao2023} and, with its convolutional neural network (CNN) architecture, is lighter than CellViT-SAM-B, which make it easier to use for training and inference.  \\

To additionally validate Histo-Miner, we performed immunohistochemistry (IHC) for \added[id=lu]{Myeloperoxidase}\deleted[id=lu]{Mypo} (\added[id=lu]{Mypo; granulocytes}), CD3 (lymphocytes), CD79a (plasma cells), CD10 (stromal cells) and  p40 (tumor cells), as well as H\&E staining on adjacent slides from the same tissue blocks. In 6 different cSCC tumors we selected 7 to 11 representative ROIs (750µm x 750µm) on H\&E-stained image and predicted the number of cells of each type using Histo-Miner (\textbf{Fig. \ref{fig4}c}). The same regions in IHC images were classified by a board certified dermatopathologist into 4 groups (- , +, ++, +++) based on the level of IHC positivity. Comparing the number of predicted cells across IHC positivity groups, we observed a significant association of the number of predicted cells per cell type and IHC positivity of the appropriate marker (\textbf{Fig. \ref{fig4}d}).

\subsection*{Tumor segmentation evaluation}
To evaluate the performance of SCC Segmenter model, we used intersection over union (IoU) between the predicted tumor regions and the groundtruth of tumor regions. The model was trained on randomly chosen 115 slides of TumSeg dataset (see \textbf{Method} section for detailed description of training procedure). The validation set contained 29 slides of the dataset. \added[id=lu]{A test set composed of 32 slides from 25 patients (patients not present in the training and validation sets) was used for evaluation. No data selection was performed to generate the test set. This test set is publicly available (see \textbf{Data availability} section)}. \deleted[id=lu]{The metrics were calculated on the validation set as there is no held-out set available and achieved segmented foreground intersection over union $\mathrm{IoU_{tissue}} = 0.800$, mean intersection over union $\mathrm{mIoU} = 0.884$, accuracy of foreground segmentation $\mathrm{Acc_{tissue}} = 0.837$, and average accuracy $\mathrm{mAcc} = 0.957$.} \added[id=lu]{The model achieves on the test set: segmented foreground intersection over union $\mathrm{IoU_{tissue}} = 0.852$, mean intersection over union $\mathrm{mIoU} = 0.907$, accuracy of foreground segmentation $\mathrm{Acc_{tissue}} = 0.892$, and average accuracy $\mathrm{mAcc} = 0.969$.}

\subsection*{Application of Histo-Miner to predict response of cSCC patients to immunotherapy} 

To demonstrate the usability of Histo-Miner in a clinical scenario, we applied it to predict outcomes of patients treated with immune checkpoint inhibitors using anti-PD1 antibodies. Immunotherapy is the most effective treatment for advanced cSCC patients, but half of all patients do not respond and reliable predictive parameters do not yet exist. \\

\begin{figure}[]
    \centering
    \includegraphics[width=0.85\textwidth]{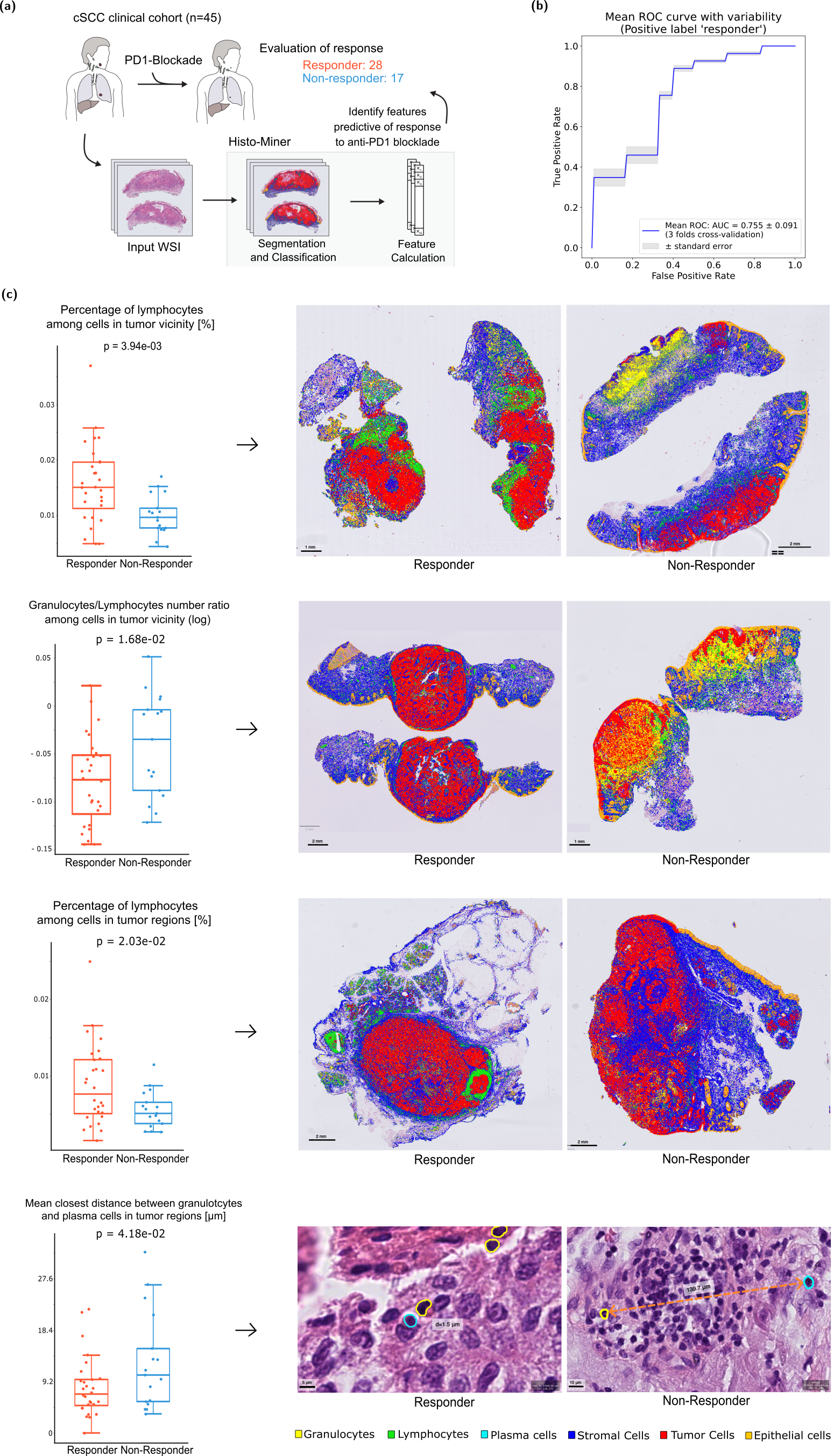}
    \caption[Caption Title]{ \normalsize  \textbf{Histo-Miner tested in a clinical scenario: prediction of CPI treatment response.}
    \small \textbf{(a)} To test the clinical utility of Histo-Miner we assembled and processed WSIs from in total n=45 cSCC patients before checkpoint inhibition (CPI) therapy.   \textbf{(b)} Mean ROC curve for the classifier keeping 19 best features and its standard error. The classifier cross-validation folds ROC curves as well as the standard deviation of the mean ROC curve are shown in \textbf{Supplementary Fig. 6}. \textbf{(c)} On the left, box plots of the 4 best features, p-value was calculated using Mann-Whitney U test. On the right, visualization of representative cases for each of the best features. For distance visualization we hide cell classes other than plasma cells and granulocytes.}
    \label{fig5}
\end{figure}

The CPI dataset consists of 45 WSIs from cSCC skin cancer of 6 medical centers before they received immune checkpoint inhibitors treatment. Each WSI is taken from a different patient. 28 of them were classified as responders (CR and PR states) and 17 patients were classified as non-responders (SD and PD states) as displayed in \textbf{Fig. \ref{fig5}a}. We performed 3 fold cross-validation with XGBoost classifier \cite{Peng2005} on the CPI dataset to predict the class of the WSIs.    \\

In each of the runs, we performed feature selection with mRMR, as described in \textbf{Methods} section, and ranked all features from most to least predictive. We then trained 107 XGBoost classifiers on the same training split, iteratively removing the least predictive feature each time, thus going from all 107 features down to the single most predictive feature. For each of these trainings we calculated balanced accuracy on the validation split. To know how many features to keep we calculated the average balanced accuracy for each training keeping $ N \in [1, 107]$ features. We kept $N_{\text{best}} = 19$ features which corresponded to the best mean balanced accuracy across all runs. We obtained mean balanced accuracy  $\bar {ba}_{N_{\text{best}}} = 0.767 \pm 0.057 $ across 3 fold cross-validation runs with the mean area under ROC curve $\bar {AUC}_{N_{\text{best}}} = 0.755 \pm 0.091 $ (\textbf{Fig. \ref{fig5}b} ). Interestingly, adding the morphology features to the set of features did not improve the classification - with $N_{\text{best}_{\text{w/morph}}}= 305$ features the balanced accuracy reaches $\bar {ba}_{N_{\text{best}_{\text{w/morph}}}} = 0.754 \pm 0.127 $. \\

\added[id=lu]{Other feature selection methods were tested but led to a worse performing feature set. We tested Boruta \cite{Kursa2010} method which, unlike mRMR that finds the smallest relevant subset of features, aims at finding all features carrying information usable for prediction. With Boruta the number of selected features varied for each fold. Classification of CPI WSIs using Boruta selected feature had a balanced accuracy of $\bar {ba}_{N_{{BORUTA}}} = \bar {ba}_{[5,0,1]} = 0.711 \pm  0.011 $ across the 3 folds. For one of the fold Boruta was enable to select features, and this for any setting of maximum depth of the random tree. In the best scenario presented here the maximum depth was set to 1, the random state was set to 0  and all other hyperparameters were kept as default values. We also ranked features following Mann-Whitney U test scores \cite{Wilcoxon1945}. We then trained 107 XGBoost classifiers on the same training split, iteratively removing the least predictive feature each time, thus going from all 107 features down to the single most predictive feature (as we did for mRMR). The best balanced accuracy was in this case $\bar {ba}_{N_{MannWhitney}} = 0.715 \pm  0.069 $ with $N_{MannWhitney} = 6$. \\} 

\added[id=lu]{Using mRMR feature selection,} the 4 most representative features in the decreasing ranking order are: the percentage of lymphocytes among all cells in tumor vicinity, the ratio between the number of granulocytes and lymphocytes in tumor vicinity, the percentage of lymphocytes among all cells in tumor region, and the mean distances between granulocytes and plasma cells in tumor regions (\textbf{Fig. \ref{fig5}c)}. We discuss the interpretation of these features in the next section. The process behind feature selection is described in the  \textbf{Method} section.  \\

\clearpage

\section*{Discussion}
WSIs contain vast amounts of detailed information and are a rich resource for cancer diagnosis and research. Automated digital pathology methods offer possibilities to efficiently process WSIs, uncovering quantitative information as well as subtle patterns and features that may be inaccessible or indiscernible to human experts. Moreover, they enable automated cell type classification and detailed description of tissue composition and architecture. These approaches have rarely been applied to non-melanoma skin cancers \added[id=lu]{like}\deleted[id=lu]{such} cSCC despite the fact that cSCC alone affects more than 1 Million individuals in the USA every year \cite{Howell2023}. A major obstacle to development of automated methods for non-melanoma skin cancer slide analysis has been the high similarity between tumor cells and non-tumor skin cells. Here, we present the Histo-Miner pipeline which provides single-cell insights into skin tumor WSIs. Our methods not only generate precise and complete information on the patient biopsy composition, we also demonstrate how this information allows to predict patient outcomes and give interpretable insights into the determinants of these outcomes. Such techniques can therefore lead to discovery of previously unknown diagnostic biomarkers, leading to improved cancer detection, diagnosis, and personalized treatment strategies. \\

Histo-Miner performs segmentation and classification of cell nuclei using a CNN, SCC Hovernet, trained on our NucSeg dataset, as well as tumor segmentation using a vision transformer, SCC Segmenter, trained on our TumSeg dataset. We compare the performance of segmentation and classification of segmented cells task to state of the art methods, such as CellViT, and show improved Panoptic Quality of our approach. Based on classification and segmentation results, Histo-Miner creates feature vectors describing tissue composition and organization. A Histo-Miner user can choose which features to calculate to best describe their WSIs. Additionally, Histo-Miner models can be trained and adapted to other cancer types. \\

Given the large size of WSIs, most commonly prediction tasks in digital pathology are performed via multiple instance learning (MIL) approaches \cite{Xu2014,Shao2024} . Via WSI tesselation and patch embedding, such approaches allow to train the model and perform the prediction on the entire WSIs directly. While it is convenient to train the prediction model end-to-end using patient-level labels that are typically available in the patient records, MIL methods offer only limited interpretability. MIL paired with attention or gradient-based mechanisms \cite{Pirovano2020,Lu2021,Li2021,Shao2021,Pisula2024} allow to disentangle which WSI regions are the most predictive, however the content of these highly-predictive regions is typically assessed in a qualitative manner. In contrast, Histo-Miner represents a feature-based approach that allows for a quantitative and systematic identification of tissue characteristics that are important for prediction. Each step of the Histo-Miner pipeline from segmentation to feature selection can be visualized allowing for inspection and interpretation of the prediction results. \\

We demonstrate the applicability of our pipeline on a cohort of 45 patients treated with immune checkpoint inhibition, which is the major treatment modality for patients with advanced cSCC \cite{Winge2023}. Even though our patient cohort was relatively small and collected across 6 medical centers, our pipeline was able to accurately predict CPI response in cross-validation experiments, highlighting its potential for clinical use cases. In addition, our feature-based approach points to the features driving the classification and thus to potential insights into the tumor biology. We identified four features the most predictive of CPI response, which included a higher percentage of lymphocytes within and in the vicinity of tumor regions in responders than non-responders. This result is in line with previous studies showing lymphocyte infiltration as a predictive marker of CPI response in cSCC \cite{Ferrarotto2021}.  Interestingly, we also observed a higher granulocyte to lymphocyte ratio and a higher distance between granulocytes and plasma cells in non-responders than responders. High neutrophil-to-lymphocyte ratio in the blood of cSCC patients has been associated with worse prognosis in general and suggested to correlated with decreased CPI response \cite{Seretis2024,Strippoli2021}, but their ratio in cSCC tissues has not been described before. Similarly, very little is known about plasma cells in cSCC tumors, especially in conjunction with granulocytes. In breast cancer however, patients not responding to CPI showed high degree of granulocytes as measured by CD15 \cite{Wang2023} positivity. In mouse models of pancreatic and squamous cell lung cancer, depletion of neutrophilic granulocytes led to reduced tumor growth \cite{Mollaoglu2018,Ng2024}. While the interplay between granulocytes and T and plasma cells in cSCC requires experimental validation in future studies,  these findings indicate that both cells of the innate (granulocytes) and the adaptive arm (T and plasma cells) of the immune system may play opposing roles in modulating CPI response in cSCC. They thereby highlight one of the strengths of our pipeline, which - in contrast to more coarse approaches - classifies immune cells into 3 subcategories and provides quantitative as well as spatial information about tumor microenvironment to identify relevant factors. It can thereby help to generate novel hypotheses for follow-up investigations. \\

\added[id=lu]{A limitation of our approach is the requirement of expert-annotated samples (tumor regions, nucleus boundaries, and cell classes) that serve as ground truth for training and testing. It may contain human errors and introduce subjective biases that the models ultimately learn to replicate. In addition, tumor cell recognition using an additional tumor region segmentation model might lead to individual cancer cells outside of the predicted tumor regions as well as non-neoplastic epithelial cells within the predicted tumor regions being misclassified.} \\

\deleted[id=jo]{A difficulty we faced in our study is the recognition between the non-neoplastic epithelial and tumor cells nuclei. The morphologies of those two cell types look indistinguishable in a skin H\&E-stained sample and only context,}  \added[id=jo]{A difficulty we faced is the distinction between non-neoplastic epithelial and tumor cell nuclei. For the published pre-trained models we tested, morphologies of those two cell types were indistinguishable in a skin H\&E-stained sample if considered in isolation. Context,} such as e.g. cell localization and neighborhood, allow to distinguish them from one another. Here the combination of two deep learning models - one for cell segmentation and classification and one for tumor region segmentation - allowed us to discriminate between the two cell classes. Further studies should focus on designing a cell classifier that is able to distinguish all 6 cell classes (granulocytes, lymphocytes, plasma cells, stromal cells, tumor cells, non-neoplastic epithelial) without using an additional tumor region classifier. Such model should incorporate a broader context including surrounding cells in a patch in the prediction process. \

\clearpage

\section*{Data availability}
\label{data_availability}
The checkpoints of SCC Hovernet and SCC Segmenter are available on Zenodo repository \href{https://zenodo.org/records/13970198?token=eyJhbGciOiJIUzUxMiJ9.eyJpZCI6IjViYmI0N2ZlLWIzZjEtNDQxZS1hMjYyLTFhNDJmYjQ3NTUxNCIsImRhdGEiOnt9LCJyYW5kb20iOiIyM2QyNzA0NWY5OGY2NjE2ZjdhMDVkM2RiMjBiNjM3YiJ9.wWqRAYBy88bOp6MBAUPgpbvY20qSfy1RxkJVn9MpSkOLncSm4eIwF7XKy9d1UAxEZ4lVid-o9smUGykukIZv_w}{link}. \\
TumSeg and NucSeg datasets are also available on Zenodo repository \href{https://zenodo.org/records/8362593?token=eyJhbGciOiJIUzUxMiJ9.eyJpZCI6ImMyNWMwNTU2LThlMDQtNDkwNS1hN2ZkLWMzYjAyNzg4ZmJlZSIsImRhdGEiOnt9LCJyYW5kb20iOiJiZjQzZThlOGJjMzEwMWJjZjk2YzNiZTY4MGI0MDhlNSJ9.NEULZTiXnprYpvbzFrQLQyKdF7ocHaXwkwup8LyIMefQB5ydyG-BasrZ9fkbfXmaoPNS0l01BrZMcwaH5DO6oA}{link}. \\
\deleted[id=lu]{Finally} The CPI image classification dataset utilized as a use-case for Histo-Miner is also available on Zenodo repository \href{https://zenodo.org/records/13986860?token=eyJhbGciOiJIUzUxMiJ9.eyJpZCI6IjUyNDM4OGNiLTEzMTQtNDhlNi05OTI1LTVhNjNlYWVlMDNkZiIsImRhdGEiOnt9LCJyYW5kb20iOiIxYmNmMzBhYmIwNDE4YmQ2N2FjNmFjNjlhODY1MWRmOSJ9.DXUEcFrVkd7-vMPw3B52s23DXq_dPWz2B7xzuZfvpOqXqPOqCphC18n9Xf_4CARmqNGFnqxMLfFIwQjTO0-yqw}{link}.  \\
\added[id=lu]{Finally, TumSeg test set, SCC Segmenter inference on test set, the list of all features from Tissue Analyser and ranking of features after cross-validation with mRMR selection are available on Zenodo repository \href{https://zenodo.org/records/17672663?token=eyJhbGciOiJIUzUxMiJ9.eyJpZCI6IjlkZGI1MmI1LWY3OTQtNGMyZS05M2I4LTllY2IzOWNlYzQzMyIsImRhdGEiOnt9LCJyYW5kb20iOiJmNzhkZmZkNmI4MTJhYTExZWIwOWQ5ZTI1OTgyNTJmYSJ9.LYZLjjfWx8RAz_JZnGcNjiUVPU4FWdiUOLqSPxnAULPhw_rRAwBPzRHaMIDH_oAc9P8p4Qx4AwVNEM2QuuECww}{link}}. \\
All the WSIs of these datasets were anonymized and cannot be used for commercial use.

\section*{Code availability}
\label{code_availability}
The Histo-Miner code is self-sufficient and user friendly as it contains explanations on how to use it that are enough for user without computer science background. In addition to the core of the code, two github submodules are used, one from MMSegmentation zoo (segmenter model) and one from Hovernet network implementation. Thanks to this approach , the Histo-Miner repository is not dependent on the changes that are operated on the original repositories. Additionally, the code is much simpler if a user wants to edit it as it keeps only functions and scripts edited in the context of Histo-Miner. \\
The github repository is available \href{https://github.com/bozeklab/Histo-Miner}{here}.\\

\section*{\added[id=lu]{Supporting Information}}

\added[id=lu]{\textbf{S1 Material.}} \\
\added[id=lu]{This supporting document contains all supplementary tables and figures cited in the main text. It includes the following sections:}
\begin{itemize}[leftmargin=*, itemsep=1pt]
  \item Poor precision in tumor and healthy epithelial detection from state-of-the-art pretrained models
  \item SCC Hovernet loss functions
  \item Hyperparameters grid search for SCC Segmenter
  \item Probability of distance overestimation
  \item List of all features from Tissue Analyser
  \item Ranking of features after cross-validation with mRMR selection
  \item ROC curves of the best-feature classifier across CV folds
  \item Stain variability for the different cohorts of TumSeg dataset
\end{itemize}
\added[id=lu]{(PDF)}

\section*{Acknowledgments}

The authors would like to thank Christian Knetschowsky for the annotations of segmentation maps of NucSeg dataset, and to thank Alfred Kirsch for the detailed corrections and extensive verifications brought to the calculation of the probability of closest distance overestimation. \deleted[id=lu]{This work was supported by the Ministry for Culture and Science (MKW) of the State of North Rhine-Westphalia [grant number 311-8.03.03.02-147635]. LS and KB were supported by the North Rhine-Westphalia return program (311-8.03.03.02-147635) and hosted by the Center for Molecular Medicine Cologne. AF was partly funded by the Deutsche Krebshilfe through a Mildred Scheel Foundation Grant (grant number 70113307). CL was partly funded through the collaborative research center grant on small cell lung cancer (CRC1399, project ID 413326622) by the German Research Foundation (DFG). JB receives funding through the collaborative research center grant on small cell lung cancer (CRC1399, project ID 413326622) and on predictability in evolution (CRC1310, project ID 325931972) by the German Research Foundation (Deutsche Forschungsgemeinschaft, DFG), a Mildred Scheel Nachwuchszentrum Grant 70113307 by the German Cancer Aid (Deutsche Krebshilfe) and the CANTAR network (NW21-062B) funded through the program "Netzwerke 2021", an initiative of the Ministry of Culture and Science of the State of Northrhine Westphalia, Germany. ODP received funding from the Bavarian Cancer Research Center (BZKF) and Deutsche Stiftung für Dermatologie. All other authors have no funding to declare.}

\section*{\added[id=lu]{Funding}}

\added[id=lu]{This work was supported by the Ministry for Culture and Science (MKW) of the State of North Rhine-Westphalia [grant number 311-8.03.03.02-147635]. LS and KB were supported by the North Rhine-Westphalia return program (311-8.03.03.02-147635) and hosted by the Center for Molecular Medicine Cologne. AF was partly funded by the Deutsche Krebshilfe through a Mildred Scheel Foundation Grant (grant number 70113307). CL was partly funded through the collaborative research center grant on small cell lung cancer (CRC1399, project ID 413326622) \added[id=jo]{and a project grant (grant ID BR 6949)} by the German Research Foundation (DFG). JB receives funding through the collaborative research center grant on small cell lung cancer (CRC1399, project ID 413326622) and on predictability in evolution (CRC1310, project ID 325931972) by the German Research Foundation (Deutsche Forschungsgemeinschaft, DFG), a Mildred Scheel Nachwuchszentrum Grant 70113307 \added[id=jo]{and project funding (grant ID 70116929)}  by the German Cancer Aid (Deutsche Krebshilfe) and the CANTAR network (NW21-062B) funded through the program "Netzwerke 2021", an initiative of the Ministry of Culture and Science of the State of Northrhine Westphalia, Germany.} \deleted[id=lu]{ODP received funding from the Bavarian Cancer Research Center (BZKF) and Deutsche Stiftung für Dermatologie} \added[id=lu]{The funders had no role in study design, data collection and analysis, decision to publish, or preparation of the manuscript. All other authors have no funding to declare.}

\section*{\replaced[id=lu]{Competing}{Conflicts of} Interests}

\added[id=lu]{I have read the journal's policy and the authors of this manuscript have the following competing interests:} JB has received research funding from Bayer AG and travel expenses from Merck KG \added[id=jo]{\& Bicycle Therapeutics and serves as a consultant for Bicycle Therapeutics} outside of the presented work.
\replaced[id=lu]{These entities}{The funders} had no role in study design, data collection and analysis, publication decisions, or manuscript preparation. ODP received personal honoraria from Merck Sharp \& Dohme and Almirall, received travel support from Kyowa Kirin, Pierre Fabre, Sanofi and Sun Pharma outside of the presented work and is member of the advisory board of Bristol Myers Squibb and Sanofi. These entities had no role in study design, data collection and analysis, publication decisions, or manuscript preparation. All other authors declare no conflicts of interests.

\section*{Ethics approval and consent to participate}

The study was performed in agreement with the Declaration of Helsinki Institutional Review Board of the University Hospital Bonn (vote number 187/16), Ethics committee of the University Hospital Cologne (votenumbers 21–1500, 20–1082 and 22–1330-retro) and institutional review board of the TU Munich (vote number 2024–363-S-CB - 1). Need for informed consent was waived for this retrospective analysis using anonymized data.

\section*{Author Contributions}

\deleted[id=lu]{LS, CL, JB, KB designed the study. CL, DH, JB provided samples and clinical data for NucSeg, TumSeg and the not-curated H\&E nucleus dataset. DH, ODP, SD, AK, ML, AF, JL provided samples and clinical data for the CPI dataset. LS, CL, DH, JB generated and cured annotations for NucSeg and TumSeg datasets. LS conceived Histo-Miner software as well as its evaluation. CL and JB conceived the immunotherapy response case study. LS conducted statistical analyses and methods for immunotherapy response case study. LS, JB, KB reviewed the manuscript.}

\noindent\textbf{Conceptualization}: Lucas Sancéré, Carina Lorenz, Johannes Brägelmann, Katarzyna Bozek. \\
\textbf{Data Curation}: Lucas Sancéré, Carina Lorenz, Doris Helbig, Johannes Brägelmann. \\
\textbf{Formal Analysis}: Lucas Sancéré.  \\
\textbf{Funding Acquisition}:  Johannes Brägelmann, Katarzyna Bozek.  \\
\textbf{Investigation}: Lucas Sancéré, Carina Lorenz. \\
\textbf{Methodology}: Lucas Sancéré. \\
\textbf{Project Administration}: Lucas Sancéré, Doris Helbig, Johannes Brägelmann, Katarzyna Bozek. \\
\textbf{Resources}:  Doris Helbig, Oana-Diana Persa, Sonja Dengler, Alexander Kreuter, Martim Laimer, Roland Lang, Anne Fröhlich, Jennifer Landsberg\\
\textbf{Software}: Lucas Sancéré.  \\
\textbf{Supervision}: Johannes Brägelmann, Katarzyna Bozek. \\
\textbf{Validation}: Lucas Sancéré. \\
\textbf{Visualization}:  Lucas Sancéré, Carina Lorenz. \\ 
\textbf{Writing – Original Draft Preparation}: Lucas Sancéré, Johannes Brägelmann, Katarzyna Bozek.  \\
\textbf{Writing – Review \& Editing}:  Lucas Sancéré, Carina Lorenz, Johannes Brägelmann, Katarzyna Bozek.

\bibliographystyle{naturemag} % We choose the "plain" reference style
\bibliography{refs} % Entries are in the refs.bib file



\section*{Supplementary material}

## Supplementary Data

**Table of Contents**

**SCC Hovernet loss functions**

The overall training loss function $L$ is a combination of loss functions as follows:

$$L = \underbrace{\lambda_{SCC_a} L_a + \lambda_{SCC_b} L_b}_{\text{HoVer Branch}} + \underbrace{\lambda_{SCC_c} L_c + \lambda_{SCC_d} L_d}_{\text{Nuclear Pixels Branch}} + \underbrace{\lambda_{SCC_e} L_e + \lambda_{SCC_f} L_f}_{\text{Nuclear Classification Branch}}$$

where $L_a$ and $L_b$ represent the regression loss with respect to the output of the HoVer branch, $L_c$ and $L_d$ represent the loss with respect to the output of the NP branch (Nuclear Pixels branch, corresponding to the Nuclear Segmentation), $L_e$ and $L_f$ represents the loss with respect to the output at the NC branch (Nuclear Classification branch), as already defined in [14]. Here all $\lambda_{SCC} = 1$.

$$L_a = \frac{1}{N} \sum_{i=1}^{N} \big(p_i(I; \mathbf{w_0}, \mathbf{w_1}) - \varGamma_i(I)\big)^2 \tag{1}$$

$$L_b = \frac{1}{m} \sum_{i \in M} \big(\nabla_x \big(p_{i,x}(I; \mathbf{w_0}, \mathbf{w_1})\big) - \nabla_x \big(\varGamma_{i,x}(I)\big)\big)^2 \tag{2}$$
$$+ \frac{1}{m} \sum_{i \in M} \big(\nabla_y \big(p_{i,y}(I; \mathbf{w_0}, \mathbf{w_1})\big) - \nabla_y \big(\varGamma_{i,y}(I)\big)\big)^2$$

$$L_c = L_e = CE = -\frac{1}{N} \sum_{i=1}^{N} \sum_{k=1}^{K} X_{i,k}(I) \log Y_{i,k}(I) \tag{3}$$

$$L_d = L_f = DICE = 1 - \frac{2 \times \sum_{i=1}^{N} \big(Y_i(l) \times X_i(l)\big) + \epsilon}{\sum_{i=1}^{N} \big(Y_i(l)\big) + \sum_{i=1}^{N} \big(X_i(l)\big) + \epsilon} \tag{4}$$

where $I$ input Image containing $N$ pixels, $p_i(I; \mathbf{w_0}, \mathbf{w_1})$ regression output of HoVer branch at pixel $i$, $w_0$ and $w_1$ 2 sets of weights. $\varGamma_i$ defines the groundtruth of the horizontal and vertical distances of nuclear pixels to their corresponding centers of mass, the horizontal and vertical components of this map are denoted $\varGamma_{i,x}$ and $\varGamma_{i,y}$ respectively (see [14] for visualization and definitions). m denotes total number of nuclear pixels within the image and M denotes the set containing all nuclear pixels. $\nabla_x$ and $\nabla_y$ denote the gradient in the horizontal x and vertical y directions respectively. Finally, $X$ denotes the branch groundtruth, $Y$ the branch prediction, K is the number of classes and $\epsilon$ is a smoothness constant that was set to $10^{-3}$.

| Hyperparameters | Value Range |
|---|---|
| cat-max-ratio | [0.75, 0.85, 0.90, 0.95, 1.0] |
| img-scale | [(2560, 640), (5120, 1280), (5120, 5120)] |
| samples-per-gpu | [4, 8] |
| pre-training-dataset | [ImageNet, Thomas2021 dataset [46]] |

**Suppl. Table 1:** Hyperparameters search for SCC Segmenter

**Hyperparameters grid search for SCC Segmenter**

On **Table.1** the different hyperparameters tested are displayed. $cat-max-ratio$ corresponds to the max area ratio that could be occupied by single category for a given crop of the input image. $img-scale$ corresponds to the resizing of the image before cropping. These resizing can randomly be modified by a factor contained in [0.5, 2], following the data augmentation pipeline from the semantic segmentation library MMSegmentation [42].

Not all combinations were fully tested, some were aborted if the training loss did not decrease in the firsts epochs or if several hyperparameters in the set were already poorly performing in previous trainings.

Set kept: {cat-max-ratio: 0.75, img-scale: (2560, 640), samples-per-gpu: 4, pre-training-dataset: ImageNet}

**Probability of distance overestimation**

In this paper, we make use of a search algorithm to estimate the average distance of the closest cell of a given class X - source class - to the closest cell of a given class Y - target class - inside the tumor regions. The principle is to define a search area (based on the tumor bounding box dimensions) centered around the nucleus of the source cell class and to check if any cells of the target class are inside. If it is the case, we calculate all the distances and keep the smallest one. If there is no cell inside the first search area, we increase the size of this area until we find at least one cell inside. Nevertheless, the closest distance found with this method is not always the actual closest distance, but sometimes it could be an overestimation. As shown in **Supplementary Fig. 1**, sometimes a cell outside the search area can be the closest but won't be taken into account.

In this supplement we show that even if such a case can occur, it is very unlikely as the number of cells increases. We simplify the problem by taking a squared search area instead of the rectangle search area that we can have in practice. The search area is based on the tumor's bounding box shape, and tumors bounding box have low aspect ratios, so taking a square search area makes it a good approximation for evaluation of error.

We consider N cells uniformly and independently distributed inside $C_2$ of radius $r_2$ and center $O$. We note $S_1$ the inscribed square in $C_2$ of side length $2r_1$. We here have $r_2 = \sqrt{2}r_1$. A representation of this model is shown in **Supplementary Fig. 2a**.

We are interested in computing the following conditional probability:

$$P(error_N) = P(B_N \mid A_N) \tag{5}$$

where:

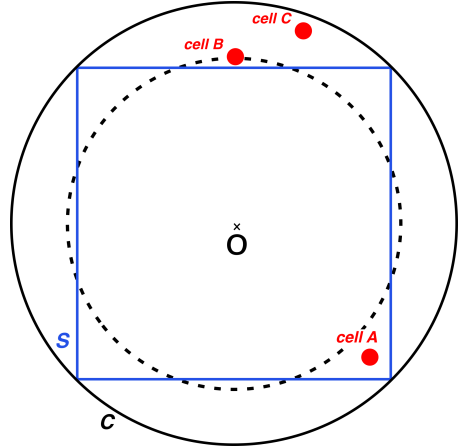

**Suppl. Fig 1: Example of distance overestimation with search area being a square.** Here we have 1 cell in the squared search area $S$ and 2 cells in $(C \setminus S)$. While calculating the minimum distance of cells from the center $O$, if we only consider cells in $S$, we will take $cellA$ as being the closest cell. In reality the closest cell is $cellB$. The distance calculated will be the distance between $cellA$ and the center, and will then be higher than the real closest distance. Nevertheless, the overestimation is bounded as expressed in **Eq. 22** with $a = 1$ as we are here in the case of a squared search area.

$$A_N : \text{(At least one cell is in } S_1\text{)}$$

$B_N$ : (There is among the cells in $(C_2 \setminus S_1)$ a cell closer to the center $O$ than all cells in $S_1$)

This probability corresponds of the probability of having an error in the distance calculation (overestimation) when at least one cell is found in the first search area $C_1$ of radius $r_1$ and center $O$. To compute this probability we first make use of the complementary event rule:

$$P(B_N \mid A_N) = 1 - P(\bar{B_N} \mid A_N) \quad (6)$$

where:

$\bar{B_N}$ : (All the cells in $(C_2 \setminus S_1)$ are further away to the center $O$ than all cells in $S_1$)

To compute $P(\bar{B_N} \mid A_N)$, we introduce for $\forall n \in [[0, N]]$:

$$U_N(n) : \text{(Exactly } n \text{ cells are in } (C_2 \setminus S_1)\text{)}$$

Using the law of total probabilities and then the definition of conditional probabilities:

$$\begin{aligned} P(\bar{B_N} \mid A_N) &= \sum_{n=0}^{N} P(\bar{B_N} \cap U_N(n) \mid A_N) \\ &= \sum_{n=0}^{N} \frac{P(\bar{B_N} \cap U_N(n) \cap A_N)}{P(A_N)} \end{aligned} \quad (7)$$

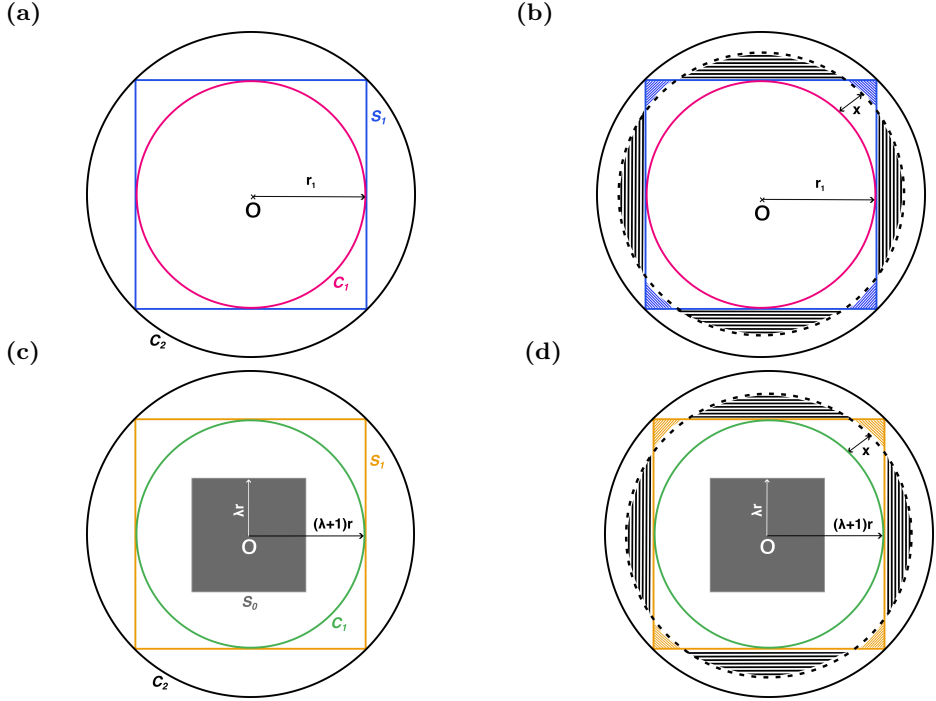

**Suppl. Fig 2: Probability of having distance overestimation representations, with search area being a square.** **(a)** During the first search, if the search box is a square of side lenght 2r, we could have a distance calculation error if a cell is in the blue square $S_1$ without being in the pink circle $C_1$ (then in $S_1 \setminus C_1$), and another cell is in black circle $C_2$ without being in in the blue square $S_1$ (then in $C_2 \setminus S_1$). Then the cell in $C_2 \setminus S_1$ could be missed from closest distance calculation and the cell in $S_1 \setminus C_1$ considered as the closest cell. In other cases, such as having at least one cell in $S_1$, no distance overestimation could be made. **(b)** To be more precise, any cell in the hatched blue area inside $S_1 \setminus C_1$ monitored by $x$ would lead for sure to an error if another cell is in the hatched black area inside $C_2 \setminus S_1$ monitored by $x$. For now we focused on the probability of overestimation in the case that at least one cell is found during the first search. If no cell is found, then the calculation differs for the further searching steps as visualized in **(c)** & **(d)**. In such cases, a specific new area cannot contain any cells, as none were found during prior searches, impacting the full calculation.

Now we note that $\forall n \in [[0, N-1]], U_N(n) \subset A_N$, so that $A_N \cap U_N(n) = U_N(n)$. We note moreover that if $n = N$, then no cells are is in the search area of $S_1$. So $U_N(N) \cap A_N = \emptyset$. Finally, noting that $U_N(0) \subset \bar{B}_N$ and using the definition of conditional probabilities, we have:

$$P(\bar{B}_N | A_N) = \frac{P(U_N(0))}{P(A_N)} + \sum_{n=1}^{N-1} \frac{P(\bar{B}_N \cap U_N(n))}{P(A_N)} \tag{8}$$

$$= \frac{1}{P(A_N)} \Big[ P(U_N(0)) + \sum_{n=1}^{N-1} P(U_N(n)) P(\bar{B}_N \mid U_N(n)) \Big]$$

As $U_N$ is a binomial trial, we can calculate $P(U_N(n))$ as follows:

$$P(U_N(n)) = \binom{N}{n} \Big( \frac{\text{area}(C_2 \setminus S_1)}{\text{area}(C_2)} \Big)^n \Big( 1 - \frac{\text{area}(C_2 \setminus S_1)}{\text{area}(C_2)} \Big)^{N-n} \tag{9}$$

$$= \binom{N}{n} \Big( 1 - \frac{2}{\pi} \Big)^n \Big( \frac{2}{\pi} \Big)^{N-n}$$

We then deduce that:
$$P(U_N(0)) = \left(\frac{2}{\pi}\right)^N \tag{10}$$

As $A_N$ : (At least one cell is in $S_1$), $\bar{A}_N$ : (No cell is in $S_1$) which is equivalent to $\bar{A}_N$ : (All N cells are in $(C_2 \setminus S_1)$), then $\bar{A}_N = U_N(N)$ and using the complementary event rule:

$$\begin{aligned}P(A_N) &= 1 - P(\bar{A}_N) \\ &= 1 - P(U_N(N)) \\ &= 1 - \left(1 - \frac{2}{\pi}\right)^N\end{aligned} \tag{11}$$

We now set two new events introducing $d = r_1 + x$ as shown in **Supplementary Fig. 2b** :

$V_N(n,d)$ : (There are $N-n$ cells in $S_1$ and the closest cell to the origin $O$ is at a distance $d$ from $O$)

And:

$\tilde{V}_N(n,d)$ : (There are $N-n$ cells in $S_1$ and they are all located at a distance greater than $d$ from $O$)

Then, using the law of total probability for continuous univariate distributions:

$$P(\bar{B}_N \mid U_N(n)) = \int_0^{\sqrt{2}r_1} P\big(\bar{B}_N \mid (U_N(n) \cap V_N(n,d))\big) \rho_n(d) \, \mathrm{d}d \tag{12}$$

where $\rho_n(d)$ is the probability density function associated to the survival function $P(\tilde{V}_N(n,d)|U_N(n))$:

$$\rho_n(d) = -\frac{\mathrm{d}P(\tilde{V}_N(n,d)|U_N(n))}{\mathrm{d}d}$$

If $d \leq r_1$ then $P(\bar{B}_N|(U_N(n) \cap V_N(n,d))) = 1$, leading to:

$$\begin{aligned}P(\bar{B}_N \mid U_N(n)) &= \int_0^{r_1} \rho_n(d) \, \mathrm{d}d + \int_{r_1}^{\sqrt{2}r_1} P(\bar{B}_N \mid (U_N(n) \cap V_N(n,d))) \rho_n(d) \, \mathrm{d}d \\ &= \left[1 - P(\tilde{V}_N(n,r_1) \mid U_N(n))\right] + \int_{r_1}^{\sqrt{2}r_1} P(\bar{B}_N \mid (U_N(n) \cap V_N(n,d))) \rho_n(d) \, \mathrm{d}d\end{aligned} \tag{13}$$

Knowing that all cells in $(S_1 \setminus C_1)$ are at a higher distance than $r_1$ from $O$, calculating $P(\tilde{V}(n,r_1)|U(n))$ gives:

$$\begin{aligned}P(\tilde{V}_N(n,r_1)|U_N(n)) &= \left(\frac{\mathrm{area}(S_1 \setminus C_1)}{\mathrm{area}(S_1)}\right)^{N-n} \\ &= \left(1 - \frac{\pi}{4}\right)^{N-n}\end{aligned} \tag{14}$$

Now we consider area($hatchblue$) and area($hatchblack$) as defined in **Supplementary Fig. 2b**. We also define the circle $C_d$ of radius $d$ ($d = r + x$).

For $d \in (r_1, r_2)$:

$$P(\tilde{V}_N(n,d) \mid U_N(n)) = \left(\frac{\text{area}(hatchblue)}{\text{area}(S_1)}\right)^{N-n} \quad (15)$$

$$= \left(\frac{\text{area}(S_1 \setminus C_d) + \text{area}(hatchblack)}{\text{area}(S_1)}\right)^{N-n}$$

$$= \left(\frac{4r_1^2 - d^2\pi + \text{area}(hatchblack)}{4r_1^2}\right)^{N-n}$$

area($hatchblack$) is the area of 4 circular segments of arc radius $d$ and sagitta ($d - r_1$) then:

$$\text{area}(hatchblack) = 4A(d - r_1, d)_{CS} \quad (16)$$

$$= 4d^2 \arccos\left(1 - \frac{d-r_1}{d}\right) - 4r_1\sqrt{d^2 - r_1^2}$$

$$= 4d^2 \arccos\left(\frac{r_1}{d}\right) - 4r_1\sqrt{d^2 - r_1^2}$$

Then:

$$P(\tilde{V}_N(n,d)|U_N(n)) = \left(\frac{4r_1^2 - d^2\pi + 4d^2 \arccos\left(\frac{r_1}{d}\right) - 4r_1\sqrt{d^2 - r_1^2}}{4r_1^2}\right)^{N-n} \quad (17)$$

$$= \left(1 + \left(\frac{d}{r_1}\right)^2 \arccos\left(\frac{r_1}{d}\right) - \frac{d^2\pi + 4r_1\sqrt{d^2 - r_1^2}}{4r_1^2}\right)^{N-n}$$

So:

$$\rho_n(d) = -\frac{\mathrm{d}P(\tilde{V}_N(n,d)|U_N(n))}{\mathrm{d}d} \quad (18)$$

$$\rho_n(d) = -(N-n)\left(\frac{1}{r_1\sqrt{1-\frac{r_1^2}{d^2}}} + \frac{2d\arccos\left(\frac{r_1}{d}\right)}{r_1^2} - \frac{\frac{4dr_1}{\sqrt{d^2-r_1^2}} + 2\pi d}{4r_1^2}\right)\left(1 + \left(\frac{d}{r_1}\right)^2 \arccos\left(\frac{r_1}{d}\right) - \frac{d^2\pi + 4r_1\sqrt{d^2-r_1^2}}{4r_1^2}\right)^{N-n-1}$$

To find $P(\bar{B}|U(n), A)$ we are left with calculating $P(\bar{B}|U(n), V(n,d))$:

$$P(\bar{B}_N \mid (U_N(n) \cap V_N(n,d))) = \left(\frac{\text{area}(C_2 \setminus S_1) - \text{area}(hatchblack)}{\text{area}(C_2 \setminus S_1)}\right)^n \quad (19)$$

$$= \left(1 - \frac{4d^2 \arccos(\frac{r_1}{d}) - 4r_1\sqrt{d^2-r_1^2}}{(2\pi-4)r_1^2}\right)^n$$

Finally:

$$\int_{r_1}^{\sqrt{2}r_1} P(\bar{B}_N \mid (U_N(n) \cap V_N(n,d))) \rho_n(d) \, \mathrm{d}d = \quad (20)$$

$$-(N-n)\int_{r_1}^{\sqrt{2}r_1} \left(\frac{1}{r_1\sqrt{1-\frac{r_1^2}{d^2}}} + \frac{2d\arccos\left(\frac{r_1}{d}\right)}{r_1^2} - \frac{\frac{4dr_1}{\sqrt{d^2-r_1^2}} + 2\pi d}{4r_1^2}\right)\left(1 + \left(\frac{d}{r_1}\right)^2 \arccos\left(\frac{r_1}{d}\right) - \frac{d^2\pi + 4r_1\sqrt{d^2-r_1^2}}{4r_1^2}\right)^{N-n-1}\left(1 - \frac{4d^2\arccos(\frac{r_1}{d}) - 4r_1\sqrt{d^2-r_1^2}}{(2\pi-4)r_1^2}\right)^n \mathrm{d}d$$

So to sum up:

$$P(error_N) = 1 - \frac{1}{P(A_N)}\Big[P\big(U_N(0)\big) + \sum_{n=1}^{N-1} P\big(U_N(n)\big)P\big(\bar{B}_N \mid U_N(n)\big)\Big] \quad (21)$$

where $P(A_N)$ is calculated in **(11)**, $P\big(U_N(0)\big)$ is calculated in **(10)**, $P\big(U_N(n)\big)$ is calculated in **(9)** and finally $P\big(\bar{B}_N \mid U_N(n)\big)$ is developed in **(13)**, **(14)**, and **(20)**.

Now we can have a numerical resolution for the probability of error $P(error_N)$ as it was decomposed fully. $N$ number of cells from 2 to 20 we can calculate $P(error_N)$ as shown in **Fig .3**.

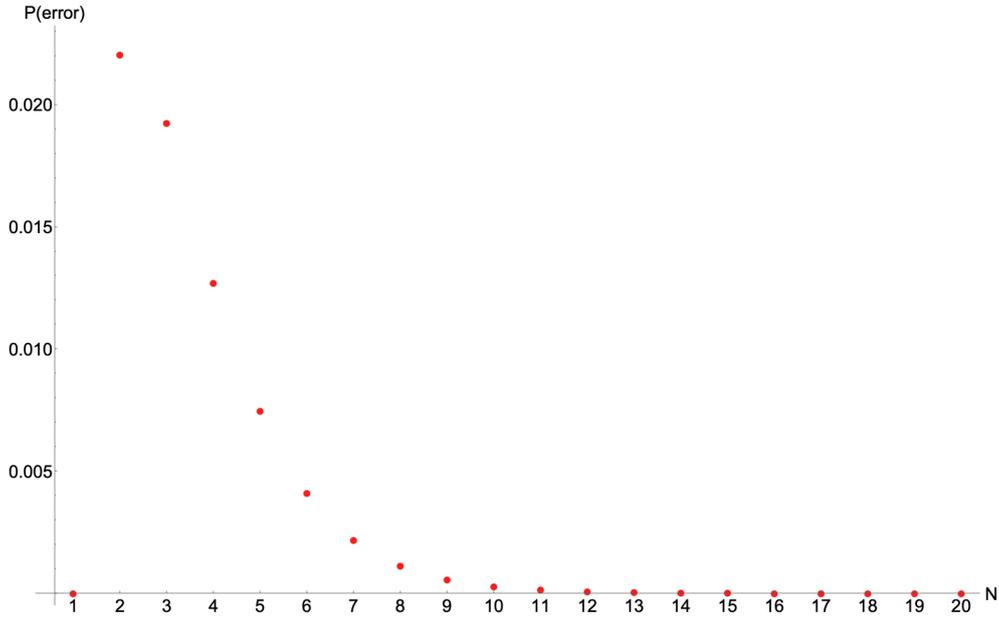

**Suppl. Fig 3: Probability of overestimation of the distance calculation in the case of a squared search area.** We focus in the case of the first search, if at least one cell is found in the first squared search area. We calculate $P(error_N)$ using the symbolic expression found in **Eq. 21**. The probability of error is 0 for $N = 1$ as expected, because to consider the wrong cell we need at least 2 cells in $C_2$. For instance we have here $P(error_{N=2}) = 0.022$, $P(error_{N=10}) = 2.828 \times 10^{-4}$ and $P(error_{N=20}) = 9.58 \times 10^{-7}$.

To validate further our probability of distance overestimation, we can verify that the calculation does not depend of the value of $r_1$ (half length of search square side). Indeed, while calculating $P(error_N)$ with Wolfram Mathematica $r_1$ is given as a variable. This simple validation is displayed in **Supplementary Fig. 4**.

We provide in our github repository (available here) a Wolfram Mathematica notebook for visualization and to calculate $P(error_N)$ for any values of $N$.

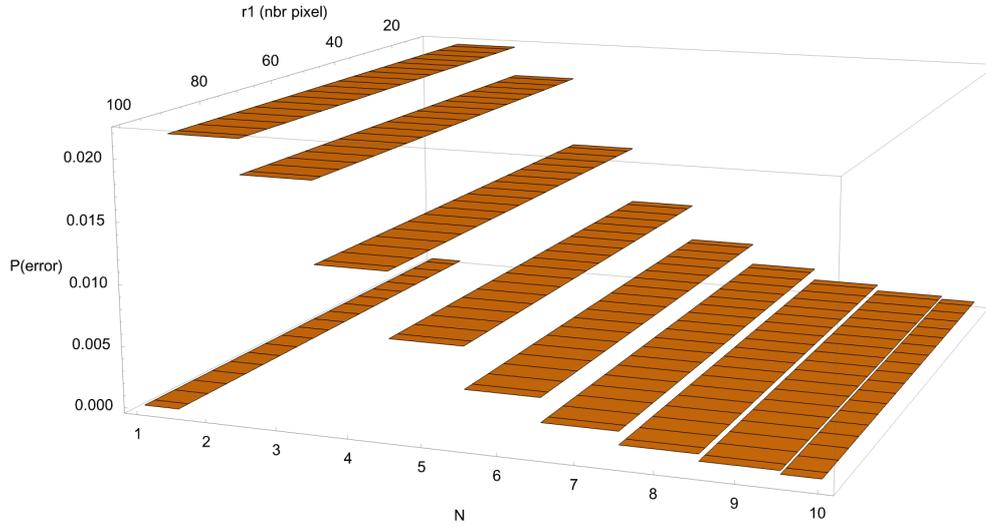

**Suppl. Fig 4: Verification of independence from $r_1$.** We verified that the probability of error is indeed independent of $r_1$ (half length of search square side) while calculating numerically the probability in Wolfram Mathematica. While checking the symbolic representation of $P(error_N)$ that was found, this independence is not easy to check, especially for $P(\bar{B}_N \mid U_N(n))$ (see **Eq. 14** & **Eq. 20**).

Finally, we can note that the relative overestimation of the distance is bounded. Going back the the case of a rectangle search area, overestimation is bounded by the aspect ratio of the tumor bounding box:

$$d_{max-overestimation} = (\sqrt{a^2 + 1} - 1)\frac{w}{2} \qquad (22)$$

where $w$ is the width of the rectangle and $a$ the aspect ratio.

We covered here the case where at least one cell is found in the first search, meaning at least one cell is in $S_1$. In practice, distances can be overestimated in subsequent search iterations following the first. Unfortunately the calculation will be slightly different. Visualization of the cases where no cell is found inside $S_1$ are available at **Supplementary Fig. 2c, 2d**.

### List of All Features from Tissue Analyser

All the features names are organized in a json file available in the CPI image classification dataset Zenodo repository here.

### Ranking of Features after Cross-Validation with mRMR Selection

The ranking of features after the 3 fold cross-validation of XGBoost with mRMR feature selection is organized in a text file available in the CPI image classification dataset Zenodo repository here.

**ROC curves of classifier with best kept features for all cross-validation folds**

In addition to figure **Fig.4.b** form the core test, we provide visualization of ROC curve for each splits.

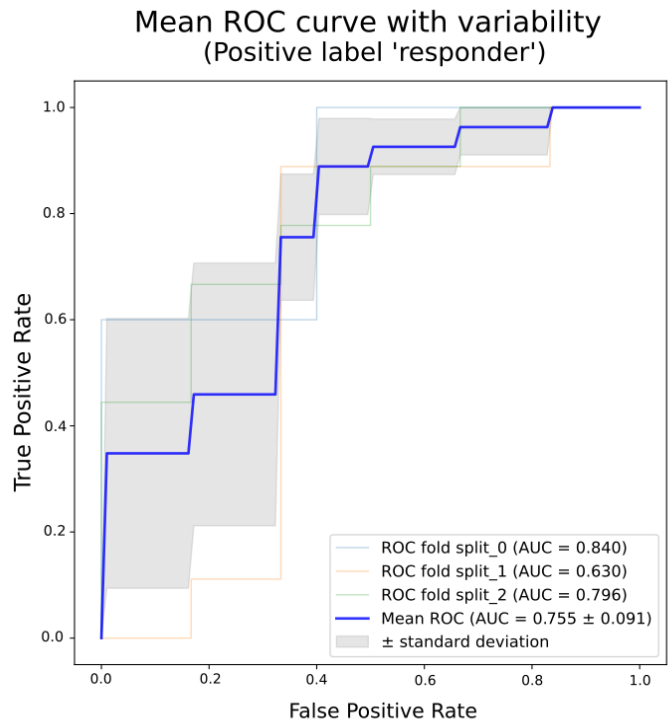

Suppl. Fig 5: Classifier cross-validation folds ROC curves.



\end{document}